\pdfoutput=1

\documentclass{article}

\PassOptionsToPackage{numbers}{natbib}

\usepackage[preprint]{neurips_2023}

\usepackage[utf8]{inputenc} 
\usepackage[T1]{fontenc}    
\usepackage[dvipsnames]{xcolor}
\usepackage[colorlinks]{hyperref}
\usepackage{url}            
\usepackage{booktabs}       
\usepackage{amsfonts}       
\usepackage{nicefrac}       
\usepackage{microtype}      
\usepackage{pifont}
\usepackage{multirow}
\usepackage{macros}
\usepackage{graphicx}
\usepackage{subcaption}
\usepackage{tabularx}
\usepackage{xspace}

\usepackage{algorithm}
\usepackage{algorithmicx}  
\usepackage[noend]{algpseudocode}
\usepackage{wrapfig}
\usepackage{listings}
\usepackage{forest}
\usepackage{adjustbox}

\hypersetup{
citecolor=NavyBlue
}

\definecolor{codegreen}{rgb}{0,0.6,0}
\definecolor{codegray}{rgb}{0.5,0.5,0.5}
\definecolor{codepurple}{rgb}{0.58,0,0.82}
\definecolor{backcolour}{rgb}{0.95,0.95,0.92}

\definecolor{tablegreen}{rgb}{0.9,0.99,0.9}
\definecolor{tablered}{rgb}{0.99,0.9,0.9}

\lstdefinestyle{mystyle}{
    backgroundcolor=\color{backcolour},   
    commentstyle=\color{codegreen},
    keywordstyle=\color{magenta},
    numberstyle=\tiny\color{codegray},
    stringstyle=\color{codepurple},
    basicstyle=\ttfamily\footnotesize,
    breakatwhitespace=false,         
    breaklines=true,                 
    captionpos=b,                    
    keepspaces=true,                 
    numbers=left,                    
    numbersep=5pt,                  
    showspaces=false,                
    showstringspaces=false,
    showtabs=false,                  
    tabsize=2
}

\newcommand{\method}{\textsc{SugarCrepe}\xspace}
\newcommand{\hncoco}{\textsc{ARO+CREPE}}
\newcommand{\replace}{\textsc{Replace}}
\newcommand{\add}{\textsc{Add}}
\newcommand{\swap}{\textsc{Swap}}
\newcommand{\negate}{\textsc{Negate}}
\newcommand{\shuffle}{\textsc{Shuffle}}

\newcommand{\replaceobj}{\textsc{Replace-Obj}}
\newcommand{\replaceatt}{\textsc{Replace-Att}}
\newcommand{\replacerel}{\textsc{Replace-Rel}}

\newcommand{\addobj}{\textsc{Add-Obj}}
\newcommand{\addatt}{\textsc{Add-Att}}

\newcommand{\swapobj}{\textsc{Swap-Obj}}
\newcommand{\swapatt}{\textsc{Swap-Att}}

\newcommand{\negclip}{\textsc{NegCLIP}\xspace}

\title{\method: Fixing Hackable Benchmarks for Vision-Language Compositionality}

\author{%
Cheng-Yu Hsieh$^{1}$\thanks{\ The authors contribute equally to this work.}, \
Jieyu Zhang$^{1}$\footnotemark[1], \
Zixian Ma$^{1}$,
Aniruddha Kembhavi$^{2}$, \
Ranjay Krishna$^{1,2}$\\
$^1$University of Washington, 
$^2$Allen Institute for Artificial Intelligence\\
\texttt{\{cydhsieh,jieyuz2,zixianma,ranjay\}@cs.washington.edu, anik@allenai.org}
}

\begin{document}

\maketitle

\begin{abstract}
In the last year alone, a surge of new benchmarks to measure \textit{compositional} understanding of vision-language models have permeated the machine learning ecosystem.
Given an image, these benchmarks probe a model's ability to identify its associated caption amongst a set of compositional distractors.
Surprisingly, we find significant biases in \textit{all} these benchmarks rendering them hackable. This hackability is so dire that blind models with no access to the image outperform state-of-the-art vision-language models.
To remedy this rampant vulnerability, we introduce \method, a new benchmark for vision-language compositionality evaluation.
We employ large language models, instead of rule-based templates used in previous benchmarks, to generate fluent and sensical hard negatives, and utilize an adversarial refinement mechanism to maximally reduce biases. We re-evaluate state-of-the-art models and recently proposed compositionality inducing strategies, and find that their improvements were hugely overestimated, suggesting that more innovation is needed in this important direction. We release \method and the code for evaluation at: \url{https://github.com/RAIVNLab/sugar-crepe}.
\end{abstract}

\section{Introduction}
\label{sec:intro}

Scholars today herald \emph{compositionality} as a fundamental presupposition characterizing both human perception and linguistic processing~\cite{cresswell1973logics}.
Through compositional reasoning, humans can comprehend new scenes and describe those scenes by composing known atoms~\cite{janssen1997compositionality,hupkes2020compositionality,bottou2014machine,chomsky1965some}.
For instance, compositionality allows people to differentiate between a photo of ``a girl in white facing a man in black'' and ``a girl in black facing a man in white''.
For a while now, vision-language research has sought to develop models that can similarly comprehend scenes and express them through compositional language~\cite{krishna2017visual,ji2020action,lu2016visual,GrundeMcLaughlin2021AGQA}.

Given its importance, a surge of new benchmarks have been proposed to evaluate whether vision-language models exhibit compositionality.
Recently,  Winoground~\cite{thrush2022winoground}, VL-CheckList~\cite{zhao2022vl}, ARO~\cite{yuksekgonul2023when}, CREPE~\cite{ma2022crepe}, and Cola~\cite{ray2023cola} have entered the machine learning zeitgeist.
Evaluation is mostly done through an image-to-text retrieval task formulation: by measuring how often models pick the description, ``a girl in white facing a man in black'' when presented with an image of it, and avoid choosing the incorrect \emph{hard negative} description, ``a girl in black facing a man in white''.

In this work, we uncover a crucial vulnerability in not just one but all these image-to-text compositionality benchmarks:
We find that a \emph{blind} model that never looks at the image, can identify the correct caption and avoid choosing the supposed ``hard negatives''.
This blind model outperforms a wide array of pretrained vision-language models across the suite of benchmarks~\cite{RadfordKHRGASAM21,ilharco_gabriel_2021_5143773,gadre2023datacomp}.
We explain this undesired hackability in existing benchmarks by showcasing that there exists a significant distributional gap between the positive and hard negative captions.
For instance, in the ARO benchmark~\cite{yuksekgonul2023when}, human-generated positive captions differ drastically from the hard negative texts generated by randomly shuffling words in the positive captions.
As new research has begun to propose methods that claim to improve compositionality on these benchmarks~\cite{yuksekgonul2023when,ray2023cola}, we find it critical to highlight our findings and propose a solution.

We propose a solution to existing hackable benchmarks by introducing \method, a new benchmark to faithfully evaluate compositionality. In curating \method, we identify two main \emph{biases}~\footnote{\ We use biases and artifacts interchangeably in the paper.} that result in the distributional gap between positive and hard negatives; and employ mechanisms to fix the shifts.
In particular, we find the current procedure in generating hard negatives introduces descriptions that are (1) not plausible and (2) non-fluent.
For example, while the caption ``olives and grapes on a plate'' is a sensical fluent caption, benchmarks often have non-plausible hard negatives like ``olives and grapes inside a plate'' or simply incomprehensible ones like ``right has word another word. There is a words'' (see Table~\ref{tab:bad_hn_examples} for more examples).
We mitigate such biases by first leveraging a modern large language model, ChatGPT~\cite{chatgpt}, to generate plausible and natural hard negative texts instead of relying on simple rule-based templates employed by existing benchmarks~\cite{ma2022crepe,yuksekgonul2023when}.
Then, we subsample the dataset through an adversarial refinement process to ensure the identified biases are maximally removed by drawing on recent dataset de-biasing work~\cite{zellers2018swag,sakaguchi2021winogrande,le2020adversarial}.
Taken together, this workflow is where \method derived its name: \textbf{S}ynthetic yet \textbf{U}nbiased \textbf{G}eneration with \textbf{A}dversarially \textbf{R}efined \textbf{C}ompositional \textbf{REP}resentation \textbf{E}valuation.
We qualitatively and quantitatively verify through both human and automatic evaluations that \method\ effectively fixes these biases.

With \method, we \emph{re}-evaluate recent methods proposed to improve compositionality. Specifically, we focus on one prominent approach that aims to improve compositionality through data augmentation. This method trains models by generating compositional hard negatives and injecting them within a training batch~\cite{doveh2023teaching,yuksekgonul2023when}.
Unfortunately, we observe that the effectiveness of this simple data augmentation approach is hugely \emph{overestimated} when evaluated on existing benchmarks, leading to limited improvements on \method.
Finally, we evaluate a wide variety of $17$ pretrained CLIP models~\cite{RadfordKHRGASAM21,ilharco_gabriel_2021_5143773,gadre2023datacomp}, and find that current models still lack compositionality.
Our results suggest that to improve compositionality, future work may need more innovative techniques.


\section{Related Work}

We situate our paper amongst existing work on vision-language compositionality, and debiasing datasets for model evaluation.

\textbf{Evaluating vision-language compositionality.}
Recent works have introduced benchmarks to evaluate the compositionality of vision-language models~\cite{RadfordKHRGASAM21}; they find that current models exhibit little compositional understanding~\cite{yuksekgonul2023when,thrush2022winoground,zhao2022vl, ma2022crepe, ray2023cola} despite their remarkable performance on downstream tasks~\cite{RadfordKHRGASAM21, 0001LXH22, singh2022flava, alayrac2022flamingo, wang2022omnivl,wang2022image, zhai2022lit}. 
Models have a hard time discerning between text containing the same words ordered differently~\cite{thrush2022winoground}. 
Models also fail to link objects to their attributes, or understand the relationship between objects~\cite{zhao2022vl, yuksekgonul2023when, ray2023cola}.
Our work finds that many of the benchmarks used to evaluate compositionality have hackable biases; blind models that do not even look at the image outperform state-of-the-art vision-language models.

\textbf{Improving vision-language compositionality.} 
To enhance vision-language models’ compositionality, new proposals suggest training strategies that utilize additional data, models, and/or losses~\cite{yuksekgonul2023when, cascantebonilla2023going, ray2023cola, doveh2023teaching, Singh2023CoarsetoFineCL}. Amongst them, one prominent approach is to explicitly train the models to distinguish hard negatives from the correct captions~\cite{yuksekgonul2023when, doveh2023teaching}.
While these approaches appear to improve compositionality on benchmarks, it is unclear if these models achieve such improvements by actually acquiring compositional understanding or by exploiting biases in these datasets. We answer this question in our evaluation.

\textbf{Debiasing dataset for faithful model evaluation.}
Several prior manuscripts have pointed out that biased datasets could lead to an overestimation of models' true capabilities~\cite{gururangan-etal-2018-annotation}. They have proposed dataset de-biasing methods to enable more faithful model evaluations~\cite{reif2023fighting,zellers2018swag,sakaguchi2021winogrande,le2020adversarial}. For instance, adversarial filtering~\cite{zellers2018swag} iteratively trains an ensemble of classifiers on different training splits and uses them to filter out ``easy'' negatives for each instance.
Building upon adversarial filtering, AFLite~\cite{sakaguchi2021winogrande,le2020adversarial} filters data instances in a more light-weight manner without retraining a model at each iteration and leads to benchmarks that more accurately represent the underlying tasks.
We use adversarial refinement to remove biases that creep into the generation of compositionality benchmarks.

\section{Current compositionality benchmarks and their biases}

A majority of existing compositionality benchmarks for vision-language models formulate the evaluation task as image-to-text retrieval~\cite{zhao2022vl,yuksekgonul2023when,ma2022crepe}. We focus on these benchmarks and discuss others~\cite{thrush2022winoground,ray2023cola} in Appendix~\ref{app:bms}.
Given an image, the model is probed to select text that correctly describes the image from a pool of candidates.
Unlike standard retrieval tasks where the negative (incorrect) candidates differ a lot from the \emph{positive} (correct) text, compositionality benchmarks intentionally design \emph{hard negative} texts that differ minimally from the positive text, in order to test whether the model understands the fine-grained atomic concepts that compose the scene.



\begin{figure}[t]
    \centering
    \includegraphics[width=\linewidth]{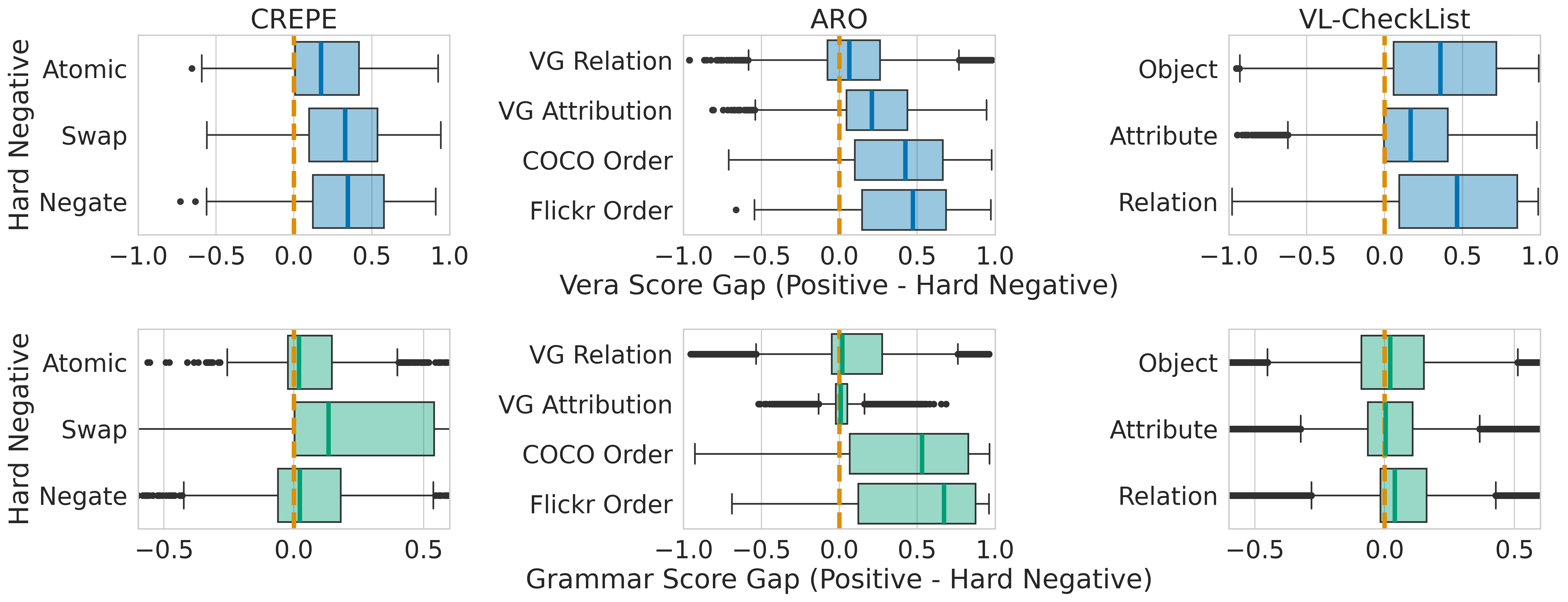}
    \caption{Top row: We define \emph{Vera score gap} as the score difference between the positive and hard negative texts: $\mathrm{Vera}(T^\mathrm{p}) - \mathrm{Vera}(T^\mathrm{n})$. The entire Vera score gap distribution lies on the positive spectrum, indicating that the template-generated hard negative texts usually have low plausibility. Bottom row: Similarly, \emph{Grammar score gap} is defined by: $\mathrm{Grammar}(T^\mathrm{p}) - \mathrm{Grammar}(T^\mathrm{n})$. On grammar score, we also find that the distribution largely rests on the positive side, suggesting that most hard negative texts in existing benchmarks exhibit grammatical errors.
    }
    \label{fig:vera_gap_existing}
\end{figure}

\textbf{Existing hard negative generation process introduces undesirable biases.}
Existing benchmarks generate hard negative texts through rule-based programmatic procedures~\cite{zhao2022vl,yuksekgonul2023when,ma2022crepe}, which produce hard negatives by replacing a word of specific type (an object, attribute, or relation) in the original text, by swapping two words, or by shuffling the word order. 
We find that such procedures introduce unintentional biases in the generated hard negatives (see Table~\ref{tab:bad_hn_examples});
specifically, we observe two major types of undesirable artifacts: (1) \emph{nonsensical} artifacts, and (2) \emph{non-fluent} artifacts.
We then utilize Vera~\cite{liu2023vera}, a plausibility estimation model, to characterize the nonsensical bias. 
To capture the non-fluent bias, we leverage a grammar-check model~\cite{morris2020textattack} that assigns high scores to grammatically correct texts.
We find that Vera and the grammar model assign higher scores to positive texts, suggesting that many hard negatives are nonsensical and not fluent (Figure~\ref{fig:vera_gap_existing}).

\begin{table*}[t]
  \centering
  \small
  \caption{Existing compositionality benchmarks rely on procedurally-generated hard negatives which often do not make logical sense or are not fluent due to grammatical errors.}
  \scalebox{0.7}{
\begin{tabular}{lll}
\toprule
    Dataset & Nonsensical Hard Negatives  & Non-fluent Hard Negatives  \\
\midrule

CREPE~\cite{ma2022crepe} & Olives and grape inside a plate. & A door with panes not in a room; the door has windows. \\
 & Ground in a basket on the flowers. & Right has word another word. There is a words. \\
 & A hair wearing a necklace, with her lady on a table. &  A shelf with books in something. There is no background. \\  
 
\midrule

ARO~\cite{yuksekgonul2023when} & The grass is eating the horse. & At brown cat a in looking a gray dog sitting is and white bathtub. \\
 & A gray bathtub is looking at a white cat. &  Scene with remarkable a ball blue a green behind chair. \\
 & Green ball with a remarkable chair behind a blue scene. & Books the looking at people are. \\  

\midrule

VL-CheckList~\cite{zhao2022vl} & Sheep is hardwood. & An man fishing a food from a wrapper using a paw at a open.\\
 & Empty zebras. & It heaving at a city. \\
 & The bush speaking in the garden. & An grouping subduing at a room access.
 
\\\bottomrule
                       
\end{tabular}
}
\label{tab:bad_hn_examples}
\vspace{-2ex}
\end{table*}

\begin{figure}[t]
    \centering
    \includegraphics[width=\linewidth]{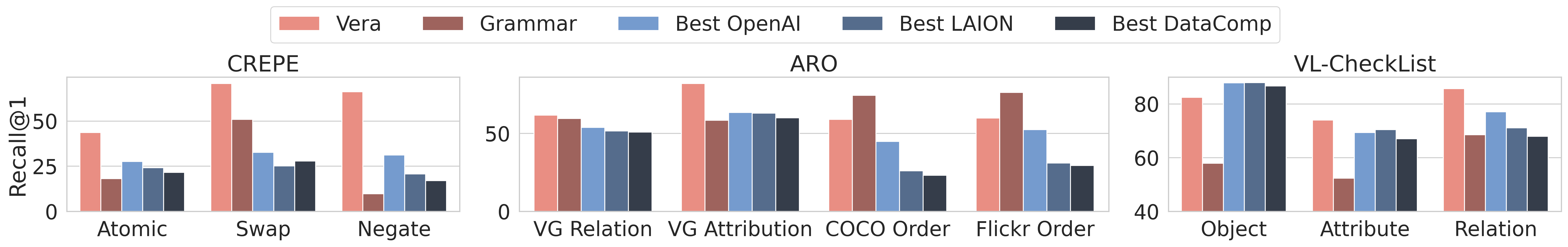}
    \caption{Blind commonsense Vera model and Grammar model outperform state-of-the-art CLIP models on nearly \emph{all} existing benchmarks by exploiting the nonsensical and non-fluent artifacts. This suggests that existing benchmarks are hackable and ineffective in measuring compositionality.
    }
    \label{fig:eval_existing}
\end{figure}

\textbf{Dataset biases render current compositionality benchmarks ineffective.}
Given the heavily-skewed score gaps, we show that blind models (\ie, Vera and the grammar model) that simply select the higher-scoring texts as positives and admittedly do not possess any vision-language compositionality, can achieve state-of-the-art performances on existing benchmarks.
We compare the the blind models against $17$ pretrained CLIP models from three sources: OpenAI in-house data~\cite{RadfordKHRGASAM21}, LAION~\cite{laion5b}, and Datacomp~\cite{gadre2023datacomp}.
We plot the performances of the blind models and the best-performing CLIP models from each category (Figure~\ref{fig:eval_existing}).
Blind models achieves state-of-the-art performances on $9$ out of $10$ existing benchmark tasks. We provide full evaluation results in Appendix~\ref{app:eval}.

\section{\method}

We introduce \method, a new benchmark for faithful evaluation of vision-language models' compositionality based on the image-text pairs of COCO~\cite{lin2014microsoft}.
\method presents two key contributions over existing benchmarks: (1) it drastically reduces the two identified dataset biases (Sec.~\ref{sec:data_generation}), and (2) it covers a broad range of fine-grained types of hard negatives (Sec.~\ref{sec:hn_types}). We present a summary comparison on compositionality benchmarks in Appendix~\ref{app:bms}.


\subsection{\method generation workflow alleviates dataset biases}
\label{sec:data_generation}

The generation procedure of \method consists of three main stages, centered around creating sensical and fluent hard negatives that close the distributional gaps to the positive texts, and ensuring a balanced distribution on the score gaps to make the final dataset robust to the identified biases.

\begin{figure}
    \centering
    \begin{minipage}{0.52\textwidth}
        \centering
        \includegraphics[width=\linewidth]{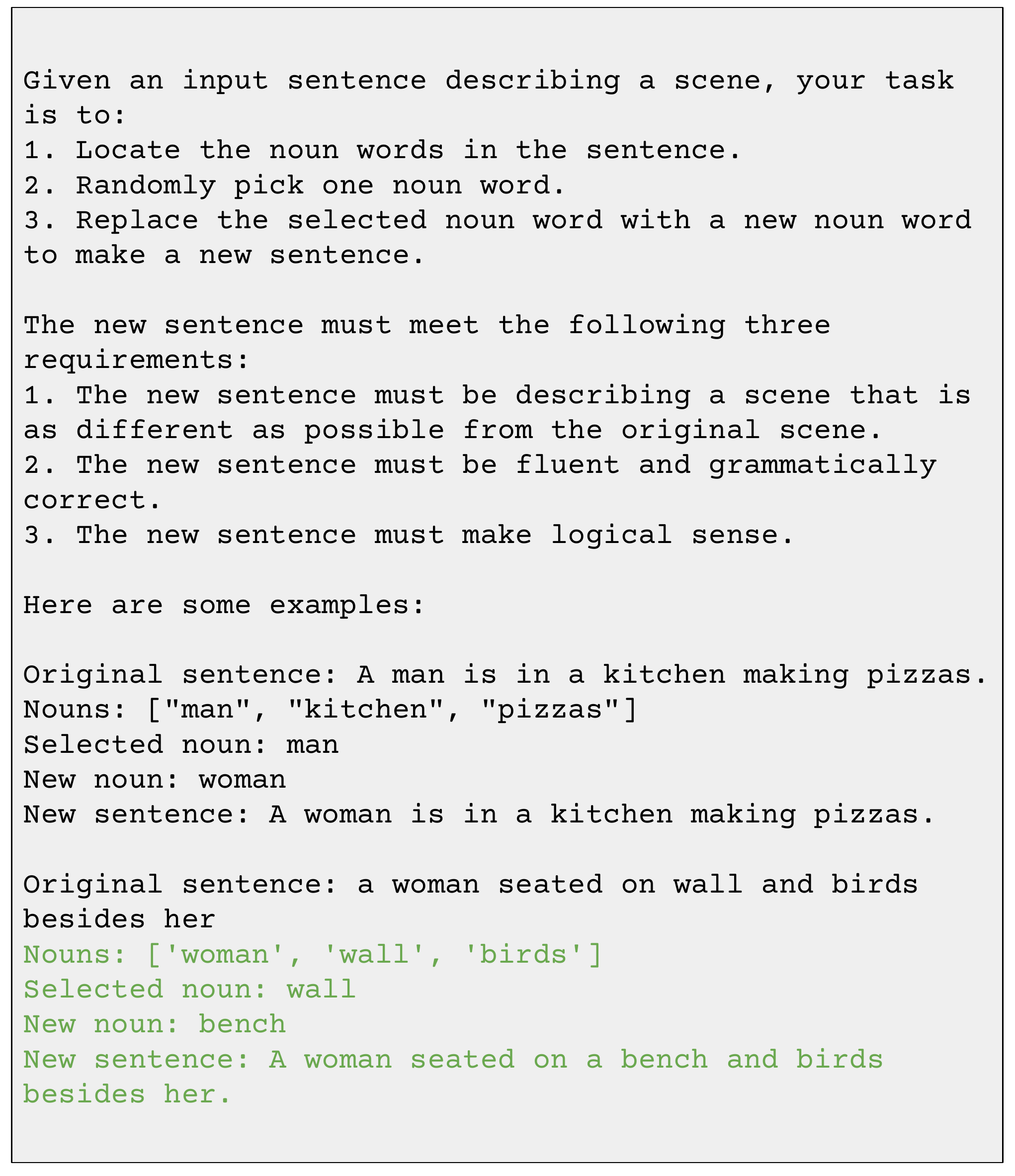}
        \caption{Example prompt (black) and actual hard negative (green) generated from  ChatGPT.}
        \label{fig:example_prompt}
    \end{minipage}\hfill
    \begin{minipage}{0.46\textwidth}
\vspace{-3mm}
\begin{algorithm}[H]  
\small
  \caption{Adversarial Refinement}
  \label{alg:ar}  
  \begin{algorithmic}[1]  
   \Require 
      Text-only model $M_1$ and $M_2$; Number of grids $K$;
      A set of candidates $\mathcal{D}=\left\{I_i, T^{\mathrm{p}}_i, T^{\mathrm{n}}_i\right\}_{i\in [N]}$, where $I_i$, $T^{\mathrm{p}}_i$, and $T^{\mathrm{n}}_i$ are $i$-th image, positive caption, and negative caption.
   \Ensure A subset $\bar{\mathcal{D}}\subset\mathcal{D}$
   \State Calculate the model score gap for each candidate $g^{(1)}_i=M_1(T^\mathrm{p}_i)-M_1(T^\mathrm{n}_i)$ and $g^{(2)}_i=M_2(T^\mathrm{p}_i)-M_2(T^\mathrm{n}_i)$
   \State Split the 2D space $[-1, 1]\times[-1, 1]$ to $K\times K$ equal-size grids.
   \State Place each candidate to a grid based on the score gaps $g^{(1)}_i$ and $g^{(2)}_i$.
   \State Initialize $\bar{\mathcal{D}}=\{\}$
    \For{each pair of grid $(G_j, G^{*}_j)$ symmetric about the original point $(0, 0)$}
        \If{$|G_j| > |G^*_j|$}
        \State Sample $|G^*_j|$ candidates from $G_j$ and put them to $\bar{\mathcal{D}}$. 
        \State Put candidates in $G^*_j$ to $\bar{\mathcal{D}}$. 
        \Else
        \State Sample $|G_j|$ candidates from $G^*_j$ and put them to $\bar{\mathcal{D}}$. 
        \State Put candidates in $G_j$ to $\bar{\mathcal{D}}$. 
        \EndIf
    \EndFor 
  \end{algorithmic}  
\end{algorithm}  
    \end{minipage}
\vspace{-2ex}
\end{figure}

\textbf{Stage 1: Generate sensical and fluent hard negatives with a large language model.}
Observing the capability of modern large language models in generating fluent and plausible texts, we leverage ChatGPT~\cite{chatgpt} to generate hard negative texts where we explicitly instruct it to avoid commonsense (logical) and fluency (grammatical) errors. To guide ChatGPT in re-writing a given positive text into its hard negative counterparts, we provide few-shot demonstrations written by the authors and leverage its in-context learning ability to generalize to unseen texts.
Figure~\ref{fig:example_prompt} shows an example demonstration used and an actual hard negative generated.
We detail all the prompt templates in Appendix~\ref{app:prompt}.
Table~\ref{tab:sugar_crepe_examples} shows the comparisons between hard negatives generated from ChatGPT in \method and that from existing benchmarks.

\textbf{Stage 2: Filter false negatives with human validation.}
A generated text is considered a valid hard negative only if it incorrectly describes the corresponding image. For example, given an image with a positive caption ``a man and a child sitting on a sofa'', a compositional change that replaces ``child'' with ``girl'' may still result in a correct caption.
To ensure the validity of the hard negatives in \method, we filter out \textit{false} negatives by manually examining the generated hard negatives and their corresponding images.

\textbf{Stage 3: De-bias dataset with adversarial refinement.}
While ChatGPT yields more sensical and fluent text, there is no guarantee that the bias between positive and negative texts is negligible.
Following dataset de-biasing work~\cite{zellers2018swag,sakaguchi2021winogrande,le2020adversarial}, we develop an adversarial refinement mechanism that maximally reduces the undesirably exploitable artifacts in \method.
Specifically, our goal is to ensure that performance improvements on \method cannot be achieved by exploiting the identified nonsensical and non-fluent biases. To accomplish this, we characterize the biases again with the commonsense and grammar models~\cite{liu2023vera,morris2020textattack}, and subsample the dataset to ensure symmetric score gap distributions on both the positive and negative sides, as shown in Figure~\ref{fig:vera_gap_ours}. 
We note the symmetry around zero implies that the commonsense and grammar scores can no longer be used to infer the ground truth positive texts.
We provide the adversarial refinement algorithm in Algorithm~\ref{alg:ar}.


\subsection{\method covers a broad range of hard negative types}
\label{sec:hn_types}

To test different aspects of vision-language models' compositional understanding, we follow CREPE~\cite{ma2022crepe} to consider various \emph{forms} of hard negatives, and follow VL-CheckList~\cite{zhao2022vl} and ARO~\cite{yuksekgonul2023when} to consider different fine-grained \emph{categories} of the atomic concepts.
In total, \method covers $7$ fine-grained types of hard negatives, as shown in Table~\ref{tab:dataset_summary}.
We introduce the dataset taxonomy below, starting from the \emph{form} of the hard negatives to its different \emph{finer-grained} variants.

\textbf{The \replace\ form.}
Given a positive text describing a scene, we generate a \replace\ hard negative by replacing an atomic concept in the original text with a new concept that makes the text mismatch with the original scene.
Based on the type of the atomic concept---object, attribute, or relation---we further categorize \replace\ hard negatives into \replaceobj, \replaceatt, and \replacerel.

\textbf{The \swap\ form.} Different from \replace, \swap\ does not introduce new concepts in the hard negatives, but a \swap\ hard negative is generated by swapping two atomic concepts of the same category in the positive text. We further categorize \swap\ into \swapobj\ and \swapatt, and omit swapping two relationships since it generally results in nonsensical texts.

\textbf{The \add\ form.} Similar to the \replace\ form, but instead of replacing an atomic concept with a new one, we generate an \add\ hard negative by adding a new atomic concept to the positive text that makes it mismatch with the original scene.
We only further categorize \add\ into \addobj~(adding object concept) and \addatt~(adding attribute concept), as adding new relationship concepts to the positive texts often make them highly implausible.


\textbf{Dataset overview.} The final evaluation set of \method consists of $7512$ examples, where the numbers for each fine-grained type are listed in Table~\ref{tab:dataset_summary}.
Each example is an image-to-text retrieval task composed of an image, a positive text, and a hard negative. On \method, random chance performance has an average accuracy of $50\%$.
We note that ARO and CREPE additionally consider \shuffle\ (randomly shuffling words in a sentence) and \negate\ (adding negation keywords ``no/not'' to a sentence) hard negatives. We however omit them in \method as \shuffle\ is very unlikely to be plausible and fluent, and \negate\ introduces irreducible keyword artifacts~\cite{ma2022crepe}.~\footnote{One can easily infer hard negatives from whether the text contains negation keywords ``no/not''.}

\begin{table*}[!t]
  \centering
  \small
  \caption{We report the number of hard negative captions of all types in \method.}
  \scalebox{0.7}{
    \begin{tabular}{lccccccc} 
    \toprule
     & \multicolumn{3}{c}{\bf \replace} & \multicolumn{2}{c}{\bf \swap} &  \multicolumn{2}{c}{\bf \add}\\ 
     \cmidrule(lr){2-4}\cmidrule(lr){5-6}\cmidrule(lr){7-8}
     
     & \textbf{Object} & \textbf{Attribute} & \textbf{Relation} &\textbf{Object} & \textbf{Attribute} & \textbf{Object} & \textbf{Attribute} \\ 
    \midrule

\textbf{\# negative captions} & 1652 & 788 & 1406  &  246 & 666  & 2062 & 692 \\

\bottomrule 
\end{tabular}
}
  
  \label{tab:dataset_summary}
\end{table*}

\section{Evaluations}
\label{sec:exp}


In this section, we qualitatively and quantitatively compare \method to existing benchmarks (Sec.~\ref{sec:exp-bias}), 
re-evaluate recent methods proposed to improve compositionality of vision-language models (Sec.~\ref{sec:exp-reeval}), and comprehensively evaluate a wide array of pretrained CLIP models (Sec.~\ref{sec:exp-eval}).



To systematically and fairly compare \method with existing benchmarks, we normalize the benchmarks by reproducing their data generation workflow using COCO~\cite{lin2014microsoft} as in \method. We utilize source code from CREPE~\cite{ma2022crepe} to generate \replace, \swap, \negate\ hard negatives and take \shuffle\ hard negatives released in ARO~\cite{yuksekgonul2023when}. We refer to this reproduced dataset as \hncoco.
In addition, we standardize the evaluation task as retrieving the correct caption from \emph{two} possible choices, \ie, a positive text and a hard negative.
This normalization sets the positive texts fixed for all benchmarks, including \method.

\subsection{\method significantly reduces dataset biases}
\label{sec:exp-bias}

\begin{table*}[!t]
  \centering
  \small
  \caption{We present example positive texts and their hard negatives in \hncoco\ (generated using existing procedures) and \method (generated with ChatGPT). \method brings significant improvements in commonsense and fluency.}
  \resizebox{\linewidth}{!}{
    \begin{tabular}{l lll} 
    \toprule
    Hard-Negative Type & Text Type & Commonsense & Fluency\\
    \midrule
    
     & Original & Two adult bears play fight in the water. & A man sitting in front of a laptop computer.\\
     \replace & \hncoco & Two adult bears play fight in the soda. & A man sitting around front of a laptop computer.\\
     & \method & A flock of ducks play fight in the water. & A man standing in front of a laptop computer.\\

     \midrule
    
    & Original & A woman standing behind a fence looking at an elephant. & Man swinging tennis racket while group of people watches. \\
     \swap & \hncoco & A fence standing behind a woman looking at an elephant. & Group swinging tennis racket while man of people watches. \\
     & \method & An elephant standing behind a fence looking at a woman. & Group of people swinging tennis racket while man watches. \\

    \midrule
    
    & Original & A teddy bear next to a stuffed fish. & A red fire hydrant on a city sidewalk. \\
     \negate\ / \add & \hncoco &  A teddy bear next to a stuffed fish. There is no teddy bear. & A red fire not hydrant on a city sidewalk.\\
     & \method & A teddy bear and a stuffed fish and a robot toy. & A red fire hydrant and a trash can on a city sidewalk. \\

\bottomrule 
\end{tabular}
}
  
  \label{tab:sugar_crepe_examples}
\end{table*}

\textbf{\method generates more sensical and fluent hard negatives.}
We validate that \method generates higher quality hard negative texts by leveraging ChatGPT than previous rule-based approaches.
Qualitatively, in Table~\ref{tab:sugar_crepe_examples}, we observe that the hard negatives in \method are more sensical and fluent compared to hard negatives in \hncoco.
We report human evaluation results in Appendix~\ref{app:sugar} that show on an average of $35\%$ of examples, hard negatives in \method have \emph{strictly} higher quality than \hncoco\ in terms of commonsense and fluency. For instance, on \swap, humans judge that \method wins $68\%$ over \hncoco\ and ties on $28\%$ of examples in terms of commonsense.
Quantitatively, in Table~\ref{tab:sugar_crepe_eval_quant_auto}, we compare the commonsense and grammar scores averaged over the hard negative texts in both \hncoco\ and \method. We see \method has much higher average scores than \hncoco. Additionally, pairwise comparisons show that \method has higher commonsense and grammar scores than \hncoco\ on $86\%$ of examples on average.

\begin{table*}[!t]
  \centering
  \small
  \caption{We compare the commonsense and grammar scores on hard negatives in \hncoco\ and \method. We report both their respective average scores and the ratio where \method has higher score than \hncoco\ in pairwise comparison. Overall, \method has hard negatives with better commonsense and grammar.}
  \scalebox{0.8}{
    \begin{tabular}{ll ccc} 
    \toprule
    & & \multicolumn{2}{c}{Average Score} \\
    \cmidrule{3-4} 
    Hard-negative Type & Metric & \hncoco & \method & Pairwise Better Ratio \\
    \midrule
    
    \multirow{2}{*}{\replace} & Commonsense & 37.46 & 50.21 & 77.71 \\
    & Grammar & 76.79 & 88.96 & 86.85\\

    \midrule
    
    \multirow{2}{*}{\swap} & Commonsense  & 23.09 & 41.57 & 78.76\\
    & Grammar & 45.67 & 80.46 & 87.02\\

    \midrule
    
    \multirow{2}{*}{\negate\ / \add} & Commonsense &  25.24 & 50.20 & 87.24\\
    & Grammar & 65.09 & 90.07 & 95.03\\

\bottomrule 

\end{tabular}
}
  
  \label{tab:sugar_crepe_eval_quant_auto}

\end{table*}

\begin{figure}[!t]
    \centering
    \includegraphics[width=\linewidth]{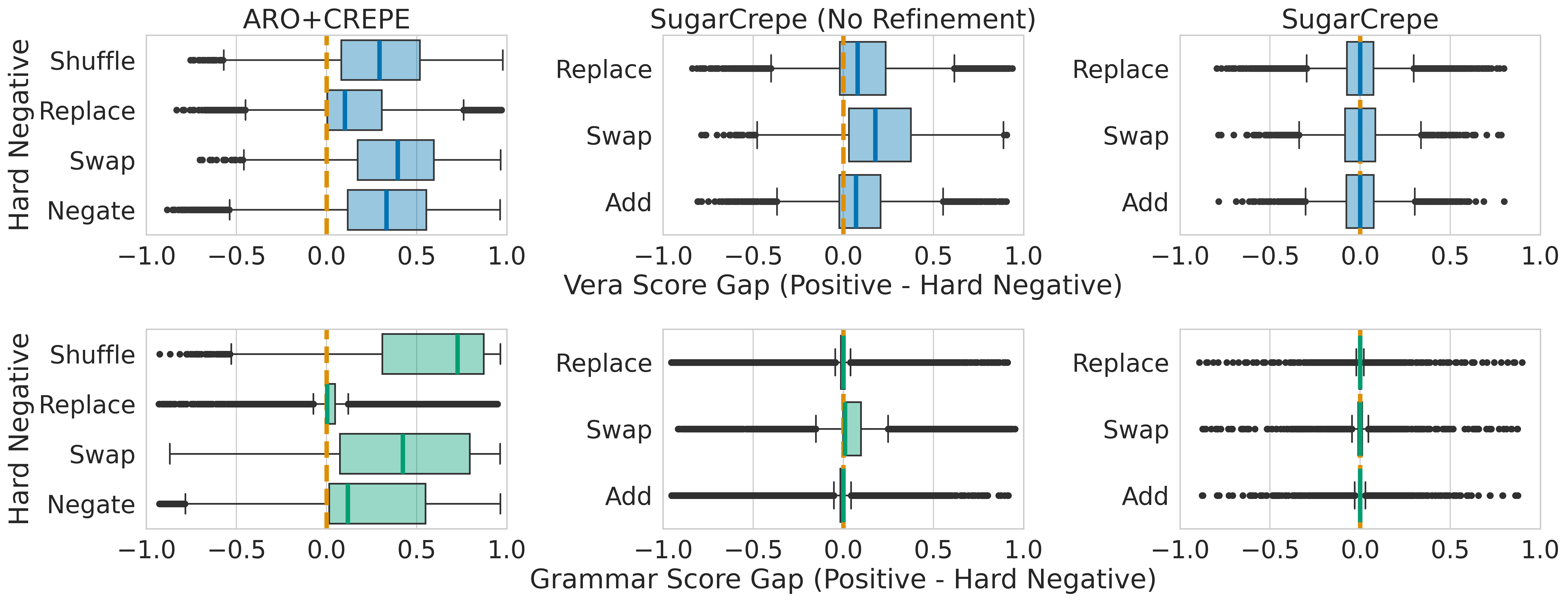}
    \caption{We compare the Vera (top row) and Grammar (bottom row) score gap distributions between \hncoco\ (leftmost column), \method without adversarial refinement (middle), and \method (rightmost). Top row: We see that Vera score gap distribution shifts from the positive spectrum to more centered around zero from \hncoco\ to \method without refinement. After adversarial refinement, we ensure the score gap distribution is centered around zero on \method. Bottom row: Similarly, from \hncoco\ to \method, we see the Grammar score gap distribution shifts from the positive spectrum to centered around zero.}
    \label{fig:vera_gap_ours}
\end{figure}

\textbf{\method disentangles the identified exploitable biases.}
We show that the final \method evaluation set maximally reduces the identified biases that could be exploited undesirably to achieve improvements on a benchmark.
Figure~\ref{fig:vera_gap_ours} visualizes the Vera/Grammar score gap distributions.
We compare the distributions between \hncoco\ and \method (before and after adversarial refinement).
First, We see that by leveraging ChatGPT, the hard negative texts in \method already have lower biases than \hncoco\ before adversarial refinement, \ie, the score gap distribution is more centered around zero.
Furthermore, we see that after adversarial refinement, the score gap distributions on the final \method evaluation set are symmetric around zero. This implies that the previously identified artifacts can no longer be exploited to infer the positive texts.
As a result, we show that the previous commonsense and grammar attacks that are extremely successful on existing benchmarks do not work on \method. As shown in Table~\ref{tab:eval_ours}, these blind models now consistently rank the \emph{last} on \method as compared to other pretrained CLIP models.

\subsection{Re-evaluating recent methods for improving compositionality}
\label{sec:exp-reeval}
Given the vulnerability of existing compositionality benchmarks, it is unclear whether recently proposed methods that show state-of-the-art performances on these benchmarks are indeed effective. Thus, we re-evaluate these methods with \method.

\textbf{Hard negative augmented training.} 
Specifically, we focus on evaluating one common \emph{data-augmentation} approach considered in~\cite{yuksekgonul2023when,doveh2023teaching}, where the core idea is to explicitly create hard negatives and train the model to distinguish them. We broadly refer to this training scheme as \negclip following \cite{yuksekgonul2023when}.
We evaluate two \negclip training schemes: finetuning and training from scratch.
For finetuning, in addition to taking the model released in \cite{yuksekgonul2023when}, we finetune another three \negclip\ models (using ViT-B/32 following~\cite{yuksekgonul2023when}) with three respective types of hard negatives (\ie, \replace, \swap, \negate) generated using CREPE's~\cite{ma2022crepe} source code.
For training from scratch, we use RN50 as the base model and train variants of \negclip by augmenting the training examples with different types of hard negatives.
We perform both training and finetuning on COCO~\cite{lin2014microsoft}.

\textbf{Improvements are overestimated due to unintentionally overfitting.}
In Table~\ref{tab:re_eval}, we first see that \negclip\ finetuned models show significant improvements on \hncoco, boosting the performance more than 10\% compared to standard CLIP finetuning on 11 out of 16 cases (highlighted in green). The lifts are especially large when the hard negative type used in finetuning matches that used in evaluation, where \negclip\ finetuned models can achieve near human-level performances. For instance, by finetuning with \replace\ hard negatives, \negclip\ reaches 94\% on \hncoco\ evaluated with \replace\ hard negatives (human performance is 95\%).
While the results on \hncoco\ suggest that \negclip\ is seemingly sufficient in equipping models with strong compositionality, we however see that the improvements brought by \negclip\ are much smaller on \method. In fact, none of the improvements on \method is larger than 10\%, and the best performing \negclip\ finetuned models still have large gaps to human-level performances, \eg, best \negclip\ model lags behind human by 23\% on \method's \swap\ hard negatives.
Similarly, when trained from scratch, we observe the same trend that \negclip's improvements are much larger on \hncoco\ than on \method. The improvements on \hncoco\ are again most pronounced when the training and testing hard negative type matches.

We attribute the stark contrast in \negclip's effectiveness on \hncoco\ and \method to model's unintentional overfitting: The \negclip\ models learned to exploit artifacts that can be used to easily distinguish hard negatives from positives on \hncoco, instead of actually improving compositionality.
Thus, when evaluated on \method where the artifacts are removed, the improvement from \negclip\ drastically reduces.
These results imply that \negclip's effectiveness is overestimated on existing benchmarks, and we may still need further innovations to fundamentally improve a model's compositionality.

\begin{table*}[h]
  \centering
  \small
  \caption{Re-evaluating hard negative augmented training shows that the method's improvements on existing benchmarks (\hncoco) are hugely overestimated, particularly when the test hard negative type matches the one used in training, which can be attributed to overfitting the artifacts.\\Color notations: \colorbox{tablegreen}{Gains compared to standard CLIP (finetuned / from scratch) $>$ 10\%}.}
  \begin{adjustbox}{max width=\textwidth}
    \begin{tabular}{lll cccc ccc} 
    \toprule
     & & & \multicolumn{4}{c}{\bf \hncoco} &  \multicolumn{3}{c}{\bf \method} \\ 
     
     \cmidrule(lr){4-7} \cmidrule(lr){8-10}
     
    \textbf{Model} & \textbf{Training} & \textbf{Hard Negative Used} & \textbf{\replace} & \textbf{\swap} & \textbf{\negate} & \textbf{\shuffle} &  \textbf{\replace} & \textbf{\swap} & \textbf{\add} \\ 
    \midrule

Human & & & 95.33 & 100 & 99.33 & 96.00 & 98.67 & 99.50 & 99.00 \\ \midrule

\multirow{5}{*}{ViT-B/32} & Pretrained & N/A & 75.71 & 71.58 & 76.89 & 72.06 & 80.76 & 63.27 & 75.09\\

& CLIP finetuned & N/A & 77.06 & 68.81 & 61.19 & 63.04 & 84.76 & 70.83 & 85.58\\
\cmidrule{2-10}

& \multirow{4}{*}{\negclip\ finetuned} & \replace & \colorbox{tablegreen}{94.51} & \colorbox{tablegreen}{90.04} & \colorbox{tablegreen}{85.06} & \colorbox{tablegreen}{88.15} & 88.27 & 74.89 & 90.16 \\
& & \swap & 82.88 & \colorbox{tablegreen}{94.48} & \colorbox{tablegreen}{77.57} & \colorbox{tablegreen}{87.00} & 85.54 & 76.21 & 86.56\\
& & \negate & 77.24 & 68.91 & \colorbox{tablegreen}{99.54} & 64.28 & 84.97 & 70.29 & 85.84 \\
&  & Released in~\cite{yuksekgonul2023when} & 85.72 & \colorbox{tablegreen}{94.35} & \colorbox{tablegreen}{83.51} & \colorbox{tablegreen}{90.45} & 85.36 & 75.33 & 87.29 \\ 

\midrule
\multirow{5}{*}{RN50}
& CLIP from scratch & N/A & 69.93 & 59.96 & 55.36 & 68.78 & 69.54 & 60.33 & 67.63\\
\cmidrule{2-10}
& \multirow{4}{*}{\negclip\ from scratch} & \replace & \colorbox{tablegreen}{89.04} & 66.51 & 60.90 & 75.23 & 74.32 & 62.65 & 72.92\\
& & \swap & 72.33 & \colorbox{tablegreen}{92.29} & 64.51 & \colorbox{tablegreen}{84.84} & 73.31 & 68.35 & 71.93\\
& & \negate & 70.09 & 60.29 & \colorbox{tablegreen}{99.45} & 69.03 & 72.74 & 60.89 & 70.47\\
& & \textsc{Rep} + \textsc{Sw} + \textsc{Neg} &  \colorbox{tablegreen}{86.30} & \colorbox{tablegreen}{88.60} & \colorbox{tablegreen}{99.34} & \colorbox{tablegreen}{82.93} & 75.26 & 67.69 & 73.08\\

\bottomrule 
\end{tabular}
\end{adjustbox}
  
  \label{tab:re_eval}

\end{table*}

\subsection{Comprehensive evaluations on existing pretrained vision-language models}
\label{sec:exp-eval}

\begin{figure}[t]
    \centering
    \includegraphics[width=\linewidth]{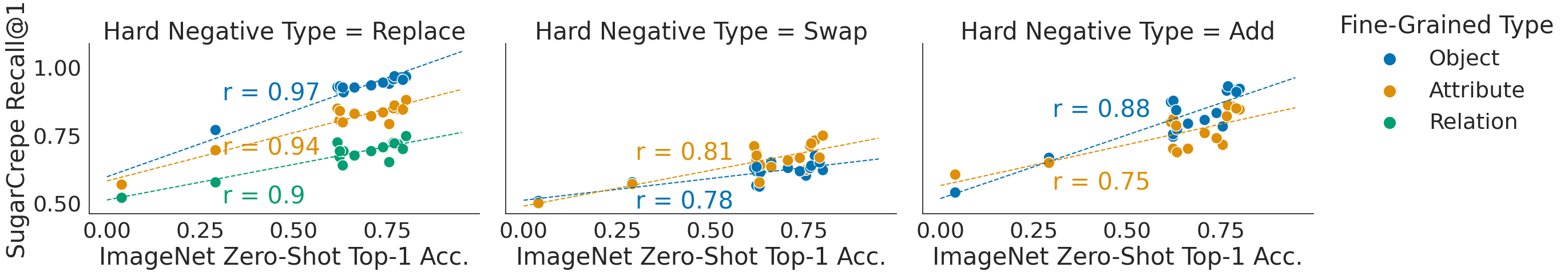}
    \caption{We plot pretrained vision-language models' zero-shot top-1 accuracy on ImageNet versus their retrieval recall@1 on \method, where $r$ is the Pearson correlation coefficient. This plot suggests that models' ImageNet zero-shot accuracy positively correlates with their compositionality.}
    \label{fig:imagenet_correlation}
\end{figure}

\begin{table*}[t]
  \centering
  \small
  \caption{Our evaluation of pretrained CLIP models on \method shows that they demonstrate compositionality on some hard negatives but are far from human performance on others, especially on \swap\ hard negatives or ones perturbing attributes and relations (also illustrated in Figure~\ref{fig:imagenet_correlation}: lower overall performance on \swap, and lower performances on attributes/relations compared to objects).}
  \scalebox{0.6}{
    \begin{tabular}{llrr ccccccc} 
    \toprule
     &&&& \multicolumn{3}{c}{\bf \replace}  &  \multicolumn{2}{c}{\bf \swap} & \multicolumn{2}{c}{\bf \add} \\ 
     \cmidrule(lr){5-7}\cmidrule(lr){8-9}\cmidrule(lr){10-11}
     
    \textbf{Source} &\textbf{Model} &\textbf{Data Size} &\textbf{Model Size (M)} & \textbf{Object} & \textbf{Attribute} & \textbf{Relation} & \textbf{Object} & \textbf{Attribute} & \textbf{Object} & \textbf{Attribute}   \\ 
    \midrule

\multicolumn{2}{c}{Human} & & & 100 & 99 & 97  & 99 & 100 & 99 & 99\\

\midrule\midrule

\multirow{2}{*}{Text-only model} &  Vera~\cite{liu2023vera} & & & 49.39 & 49.62 & 49.36 & 49.19 & 49.40 &  49.42 & 49.57\\

& Grammar~\cite{morris2020textattack} & & & 50.00 & 50.00 & 50.00 & 50.00 & 50.00 & 50.00 & 50.00 \\

\midrule\midrule

\multirow{7}{*}{OpenAI~\cite{RadfordKHRGASAM21}}    
& RN50& \multirow{7}{*}{400M} & 102& 91.77& 80.58& 69.99& 61.79& 68.47& 74.54& 69.65\\
& RN101& & 120& 92.49& 83.88& 67.07& 56.50& 65.92& 75.46& 70.09\\
& ViT-B-32& & 151& 90.92& 80.08& 69.20& 61.38& 63.96& 77.21& 68.79\\

& RN50x4& & 178& 92.68& 82.99& 67.57& 65.04& 63.36& 79.34& 70.09\\
& RN50x16& & 291& 93.46& 82.11& 69.20& 63.01& 65.77& 80.70& 75.87\\
& ViT-L-14& & 428& 94.07& 79.19& 65.15& 60.16& 62.31& 78.32& 71.53\\
& RN50x64& & 623& 94.49& 83.50& 70.63& 61.79& 66.67& 83.27& 73.99\\

\midrule

\multirow{6}{*}{LAION~\cite{laion5b}}   
& roberta-ViT-B-32& \multirow{4}{*}{2B}& 212& 92.86& 84.90& 72.40& 63.01& 71.02& 87.34& 79.91\\
& ViT-H-14&  & 986& 96.49& 84.77& 71.76& 67.48& 73.12& 92.05& 85.84\\
& ViT-g-14& & 1367& 95.76& 85.03& 72.40& 63.01& 71.17& 91.51& 82.08\\
& ViT-bigG-14& & 2540& 96.67& 88.07& 74.75& 62.20& 74.92& 92.19& 84.54\\ \cline{3-3}
& xlm-roberta-base-ViT-B-32& \multirow{2}{*}{5B}& 366& 93.16& 84.01& 69.20& 63.41& 67.57& 87.78& 81.07\\
& xlm-roberta-large-ViT-H-14& & 1193& 96.85& 86.04& 72.05& 63.82& 72.07& 93.11& 86.13\\

\midrule

\multirow{4}{*}{DataComp~\cite{gadre2023datacomp}}   
& \texttt{small:}ViT-B-32& 13M& 151& 56.90& 56.85& 51.99& 50.81& 50.00& 53.93& 60.55\\
& \texttt{medium:}ViT-B-32& 128M& 151 & 77.00& 69.54& 57.68& 57.72& 57.06& 66.73& 64.88\\
& \texttt{large:}ViT-B-16& 1B& 150& 92.68& 79.82& 63.94& 56.10& 57.66& 84.34& 78.61\\
& \texttt{xlarge:}ViT-L-14& 13B& 428& 95.52& 84.52& 69.99& 65.04& 66.82& 91.03& 84.97\\

\bottomrule 
\end{tabular}
}
  
  \label{tab:eval_ours}

\end{table*}
We present four key findings in our evaluation over $17$ pretrained CLIP models on \method, with results reported in Table~\ref{tab:eval_ours} and visualized in Figure~\ref{fig:imagenet_correlation}.

\textbf{The best pretrained CLIP models demonstrate some compositional understanding but still have overall large rooms for improvements.} Table~\ref{tab:eval_ours} shows that the largest pretrained CLIP models, \eg, OpenAI's RN50x64, LAION's xlm-roberta-large-ViT-H-14, and DataComp's ViT-L-14, achieve near-human performance on \replaceobj. However, on \replaceobj, smaller models pretrained on small datasets still suffer from big drops in performance --- 23\% and 43\% respectively for DataComp's small and medium models --- compared to humans. Additionally, on nearly all other hard negative types, there are clear gaps (larger than 10\%) between the best model performances and human performances, showing an overall large room for improvements in current models' compositionality.

\textbf{All models struggle at identifying \swap\ hard negatives, regardless of their pertaining dataset and model size.} Among the three types of hard negatives, \swap\ hard negatives present the biggest challenge to the pretrained CLIP models, even though humans can easily tell them apart from the positive captions.  We observe in Table~\ref{tab:eval_ours} that all models demonstrate low performance on both \swapobj\ and \swapatt\ hard negatives regardless of their pretraining dataset and model sizes, with the difference from human performance reaching from 27\% to 50\%.

\textbf{Existing models are object-centric, struggling to compose attributes and relations.} We find that existing pretrained models are a lot better at composing objects than attributes or relations (Table~\ref{tab:eval_ours}). This finding holds for both \replace\ and \add\ hard negatives but not the most difficult \swap\ negatives, where models perform equally poorly on both \swapobj\ and \swapatt.
On \replace\ hard negatives, even though most models achieve human-level performance on \replaceobj, they all suffer from a drop in performance on \replaceatt\ and \replacerel, where the drop is as large as 15\% and 29\% respectively.
Similarly, on \add\ hard negatives, all models except for DataComp's small:ViT-B-32 experience a decrease in performance from \addobj\ to \addatt, with the largest difference reaching 10\%. 

\textbf{Models' performance on \method correlates with their ImageNet zero-shot accuracy.}
We show in Figure~\ref{fig:imagenet_correlation} that there is a positive correlation between models' performance on \method and their zero-shot accuracy on ImageNet. This correlation is moderate on \swapobj\ and \addatt\ (Pearson correlation coefficient $r = 0.78$ and $r = 0.75$ respectively) and strong on all other hard negatives ($r > 0.8$).

\section{Conclusion}

Our investigation reveals significant biases present in existing benchmarks for the compositional comprehension capability of vision-language models. The severity of this vulnerability is exemplified by text-only models without access to the image outperforming vision-language models. To address this, we introduce \method, a novel benchmark for evaluating the compositionality of vision-language understanding. Unlike previous benchmarks that relied on rule-based templates, we leverage large language models to generate less biased negatives and employ adversarial filtering mechanisms to minimize biases. Through reassessment of state-of-the-art models and recently proposed compositionality inducing mechanisms, we uncover a significant overestimation of their advancements, underscoring the need for further innovation.

\clearpage 

\bibliographystyle{plain}
\bibliography{refs}

\clearpage

\appendix
\section{Limitation, future work, and societal impact}

\subsection{Limitation and future work}
There are several limitations to this work that future research can further explore. First, we focus our scope on compositionality benchmarks formulated as image-to-text retrieval task. While this is currently the most prevailing evaluation framework, future research can characterize compositionality evaluation as text-to-image retrieval problem, as in the initial efforts considered by~\cite{ray2023cola,thrush2022winoground}. More importantly, we hope our work can guide future efforts in creating and ensuring faithful compositionality benchmarks in text-to-image form.
Second, in this work, we identify \emph{two} human interpretable dataset biases, the nonsensical and non-fluent biases, which may not cover all dataset artifacts that could possibly be exploited by a model. Future work may utilize more sophisticated techniques to remove spurious dataset artifacts beyond human comprehension~\cite{le2020adversarial}.
Finally, we focus our evaluations on contrastively learned vision-language models~\cite{RadfordKHRGASAM21}. Future work should include and characterize the compositionality of modern generative vision-language models~\cite{alayrac2022flamingo,chen2022pali,li2023blip2,tschannen2023image}.

\subsection{Societal impact}
As vision-language models such as CLIP~\cite{RadfordKHRGASAM21} are becoming the foundation models for many downstream applications~\cite{rombach2022high,ramesh2022hierarchical}, it is imperative to understand the limitations of these models to avoid misuses and undesirable outcomes~\cite{cho2022dall,bianchi2022easily}. Compositionality benchmarks probe a model's understanding of finer-grained concepts, and hence allow us to identify blind spots~\cite{yuksekgonul2023when,zhao2022vl,ma2022crepe} of seemingly powerful models deemed by standard classification and retrieval benchmarks~\cite{Deng2009ImageNetAL,lin2014microsoft}.
Our work further alleviates common artifacts in existing compositionality benchmarks that result in overestimation of a model's capability. We hope our proposed benchmark \method leads to more faithful assessment of a vision-language model's compositionality, and can hence guide more accurate usages of the models.
Nevertheless, we note that strong performances on \method do not imply perfect models. We envision \method being one of the many benchmarks used to comprehensively understand the abilities of vision-language models from various aspects.


\section{Implementation details}

\subsection{Hardware information}
All experiments are run on a machine with an Intel(R) Xeon(R) CPU E5-2678 v3 with a 512G memory and two 48G NVIDIA RTX A6000 GPUs.

\subsection{Dataset sources}
We obtain all existing datasets from their original sources released by the authors.
We refer readers to these sources for the dataset licenses. To the best of our knowledge, the data we use does not contain personally identifiable information or offensive content.

\begin{itemize}
    \item CREPE~\cite{ma2022crepe}: We obtain CREPE dataset from its official repository~\footnote{\url{https://github.com/RAIVNLab/CREPE}}.

    \item ARO~\cite{yuksekgonul2023when}: We obtain ARO dataset from its official repository~\footnote{\url{https://github.com/mertyg/vision-language-models-are-bows}}.

    \item VL-CheckList~\cite{zhao2022vl}: We obtain VL-CheckList dataset from its official repository~\footnote{\url{https://github.com/om-ai-lab/VL-CheckList}}.

    \item COCO~\cite{lin2014microsoft}: We obtain COCO from its official project website~\footnote{\url{https://cocodataset.org/}}.
\end{itemize}

\subsection{Software configuration}

\textbf{Models.} We detail the sources of the pretrained models we use in the paper, and the hyper-parameters used in training our own models.

\begin{itemize}
    \item Vera model~\cite{liu2023vera}: We obtain pretrained Vera model released by its author~\footnote{\url{https://huggingface.co/liujch1998/vera}}.
    \item Grammar model~\cite{morris2020textattack}: We obtain the Grammar model released by the authors~\footnote{\url{https://huggingface.co/textattack/distilbert-base-uncased-CoLA}}.
    \item All pretrained CLIP models: We obtain all pretrained CLIP models' weights from OpenCLIP~\footnote{\url{https://github.com/mlfoundations/open_clip}}.
    \item \negclip\ models: We obtain weights for pretrained \negclip released by the authors~\footnote{\url{https://github.com/mertyg/vision-language-models-are-bows}}. For training from scratch and finetuning, we train RN50 and ViT-B/32 based on OpenCLIP codebase and set hyper-parameters as the following: number of warmup steps is 1000, batch size is 256, learning rate is 1e-4, weight decay is 0.1, number of epochs is 30. We augment the original CLIP loss with hard negative captions following \negclip~\cite{yuksekgonul2023when}.
\end{itemize}

\textbf{Evaluations.}
We base our evaluation framework on OpenCLIP~\cite{ilharco_gabriel_2021_5143773}. We follow all default hyper-parameters used for evaluating models.

\section{Vision-language compositionality benchmarks}
\label{app:bms}

We provide an overview of existing vision-language compositionality benchmarks below, with Table~\ref{tab:benchmark_comp} summarizing the dataset comparisons.

\subsection{Image-to-text formulation}
A majority of current benchmarks formulate the evaluation task as image-to-text retrieval problem. These benchmarks generate hard negative texts procedurally through rule-based templates, where each benchmark considers different types of hard negatives.

\textbf{VL-Checklist~\cite{zhao2022vl}.} VL-CheckList aims at evaluating vision-language models' understanding of different objects, attributes, and relationships. It contains \replace\ hard negatives generated by replacing atomic parts of the positive texts with other foils.
VL-CheckList further breaks the hard negatives down into more granular categories based on the type of the replaced atomic part, \ie, object, attribute, or relationship.

\textbf{ARO~\cite{yuksekgonul2023when}.} ARO focuses on models' understanding of different relationships, attributes, and order information. It considers \swap\ and \shuffle\ hard negatives. \swap\ hard negatives are generated by swapping two words in the positive texts; on the other hand, \shuffle\ hard negatives are generated by shuffling words in the positive texts. ARO further divides \swap\ hard negatives into attribute or relationship type.

\textbf{CREPE~\cite{ma2022crepe}.} CREPE is a large-scale evaluation benchmark that includes three types of hard negatives: \replace, \swap\ and \negate. \replace\ and \swap\ hard negatives are generated as in VL-CheckList and ARO. In addition, \negate\ hard negatives are generated by adding negation keywords (\ie, \textit{not} or \textit{no}) to the original positive texts. The hard negatives are not further divided into fine-grained types (object, attribute, or relations).

\subsection{Text-to-image formulation}
Complementary to image-to-text formulation, compositionality can as well be evaluated by probing a model to select an image that best matches a given text description, against other hard negative images as distractors. Unlike hard negative texts, hard negative images are more difficult to obtain and thus current text-to-image compositionality benchmarks are smaller at scale.

\paragraph{Winoground~\cite{thrush2022winoground}.} Winoground is a small dataset manually curated by human annotators. Each example in the dataset contains two images and two matching captions, where both captions contain identical words that appear in different orders. Note that Winoground can be used for either image-to-text or text-to-image retrieval. While the original intention for Winoground is to evaluate vision-language compositionality, recent work~\cite{diwan2022winoground} has pointed out that solving the tasks in Winoground requires not just compositional vision-language understanding, but additionally a suite of other abilities such as commonsense reasoning, or distinguishing visually difficult images.

\paragraph{Cola~\cite{ray2023cola}.} Cola tests a vision-language model's ability to select an image that correctly matches a given caption, against another distractor image with the same objects and attributes but in the wrong composition. The image pairs are mined from existing datasets. As a result, the final evaluation set is relatively small in size ($210$ examples in total).

We deem text-to-image evaluation as important as image-to-text evaluation. Future work can explore approaches to generate or mine compositional hard negative images at scale, as preliminarily explored in \cite{ray2023cola,yuksekgonul2023when}.

\begin{table}[t]
 \caption{Summary on vision-language compositionality benchmarks. \method considers image-to-text formulation to enable larger scale evaluation set. In addition, \method considers a wide range of hard negative types. \shuffle\ and \negate\ are omitted as they introduce inevitable biases discussed in Sec.~\ref{sec:hn_types}.}
 \centering
\scalebox{0.8}{
\begin{tabular}{lccccccc}
\toprule 
&&& \multicolumn{5}{c}{\bf Hard Negative Text Type}  \\ 
     \cmidrule{4-8}
{\textbf{Benchmark}} & {\textbf{Task Formulation}}  &  {\textbf{Scale}} & {\textbf{\shuffle}} & {\textbf{\replace}}  & {\textbf{\swap}}  & {\textbf{\negate}} & {\textbf{\add}}\\
\midrule

VL-CheckList~\cite{zhao2022vl} & Image-to-Text & $>1000$ & & \checkmark\\
\midrule
ARO~\cite{yuksekgonul2023when} & Image-to-Text & $>1000$ & \checkmark & & \checkmark\\ 
\midrule
CREPE~\cite{ma2022crepe}  & Image-to-Text & $>1000$ & & \checkmark & \checkmark & \checkmark\\
\midrule
Winoground~\cite{thrush2022winoground} & Image-to-Text / Text-to-Image & $400$ & & & \checkmark\\ 
\midrule
Cola~\cite{ray2023cola} & Text-to-Image & $210$ &  &  &  N/A \\

   \midrule
   \midrule
\method  & Image-to-Text & $>1000$ & & \checkmark & \checkmark &  & \checkmark

   \\ \bottomrule

\end{tabular}
}
\label{tab:benchmark_comp}
\end{table}



\section{\method~}
\subsection{Taxonomy}
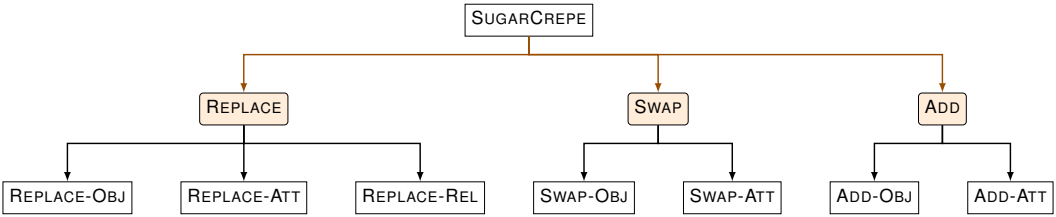
\begin{figure}[h]
\begin{adjustbox}{max width=\linewidth}
\begin{forest}
  for tree={
    font=\sffamily,
    draw,
    align=center,
    edge={thick, -latex},
    l sep+=20pt,
    s sep+=20pt,
    anchor=center,
    child anchor=north,
    parent anchor=south,
    edge path={
      \noexpand\path [draw, \forestoption{edge}] (!u.parent anchor) -- +(0,-10pt) -| (.child anchor)\forestoption{edge label};
    },
    if level=1{
      edge path={
        \noexpand\path [draw, \forestoption{edge}] (!u.parent anchor) -- +(0,-10pt) -| (.child anchor)\forestoption{edge label};
      },
      rounded corners=2pt,
      fill=orange!15,
      edge={thick, orange!60!black, -latex},
    }{},
  }
  [\method, align=center, anchor=east
    [\replace
      [\replaceobj]
      [\replaceatt]
      [\replacerel]
    ]
    [\swap
      [\swapobj]
      [\swapatt]
    ]
    [\add
      [\addobj]
      [\addatt]
    ]
  ]
\end{forest}
\end{adjustbox}
\caption{Taxonomy of hard negatives considered in \method.}
    \label{fig:sugarcrepe-tax}
\end{figure}

Figure~\ref{fig:sugarcrepe-tax} shows the taxonomy of \method. We first categorize the hard negatives based on their forms: \replace, \swap, and \add. We then further divide each type of hard negatives into finer-grained sub-categories based on the type (object, attribute, or relation) of the atomic concept altered. \method covers a total of $7$ fine-graind hard negative types.

\subsection{Hard negative generation procedure and templates}
\label{app:prompt}

To generate hard negatives in \method, we come up with three different prompt templates for the three hard negative types considered: \replace, \swap, and \add. Each template consists of task instruction for generating the corresponding type of hard negatives and several ($7$ or more) few-shot demonstrations. We describe the general generation procedure and example prompt templates below and refer readers to our dataset repository for the full prompts used~\footnote{\url{https://github.com/RAIVNLab/sugar-crepe}} .

\begin{figure}[t]
\centering
\begin{subfigure}{.4\textwidth}
  \centering
  \includegraphics[width=\linewidth]{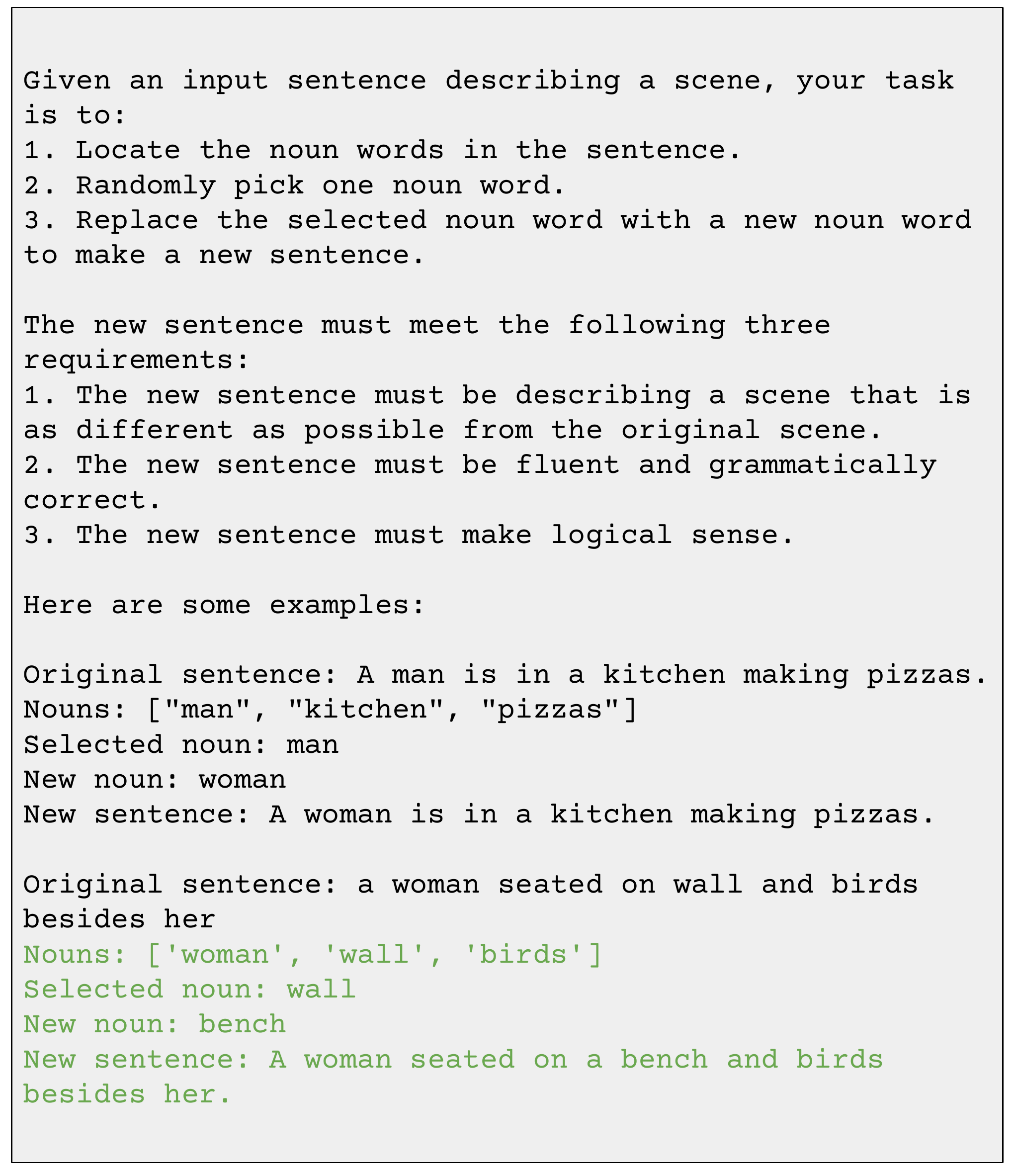}
  \caption{\replaceobj.}
\end{subfigure}
\begin{subfigure}{.4\textwidth}
  \centering
  \includegraphics[width=\linewidth]{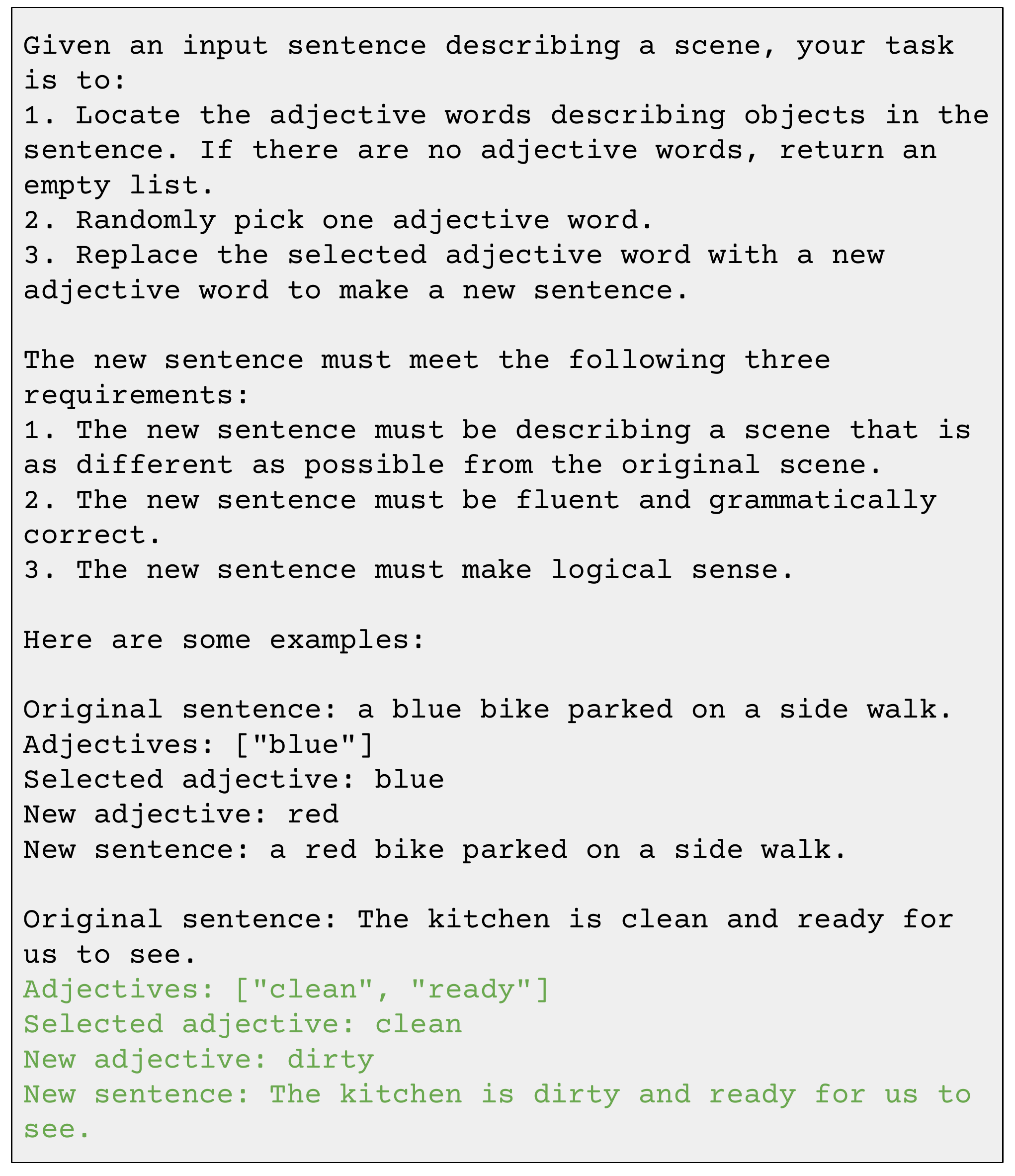}
  \caption{\replaceatt.}
\end{subfigure}
\begin{subfigure}{.4\textwidth}
  \centering
  \includegraphics[width=\linewidth]{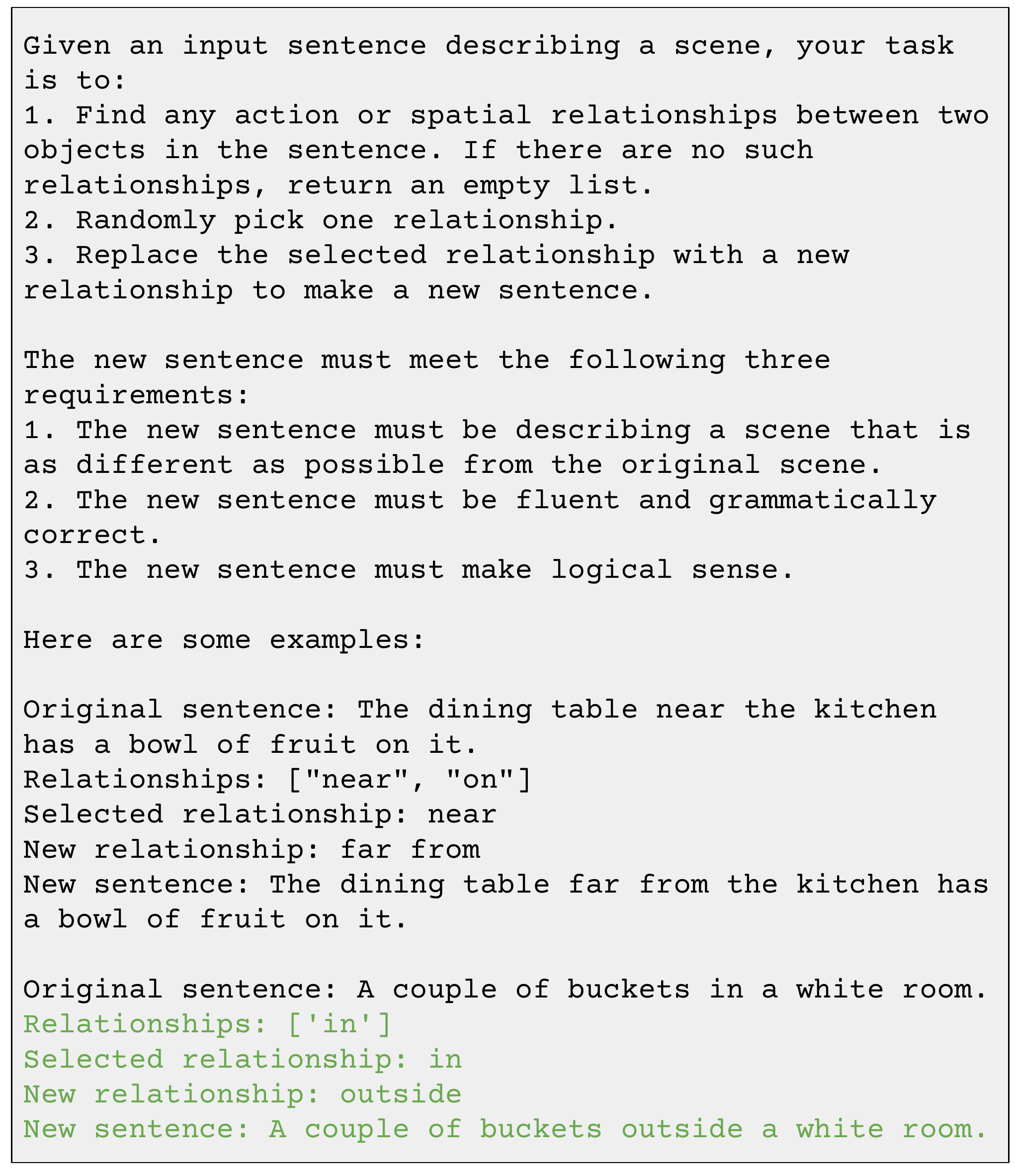}
  \caption{\replacerel.}
\end{subfigure}
\caption{Example prompt templates (black) and outputs (green) from ChatGPT for \replace\ hard negatives.}
\label{fig:prompt-replace}
\end{figure}

\begin{figure}[t]
\centering
\begin{subfigure}{.4\textwidth}
  \centering
  \includegraphics[width=\linewidth]{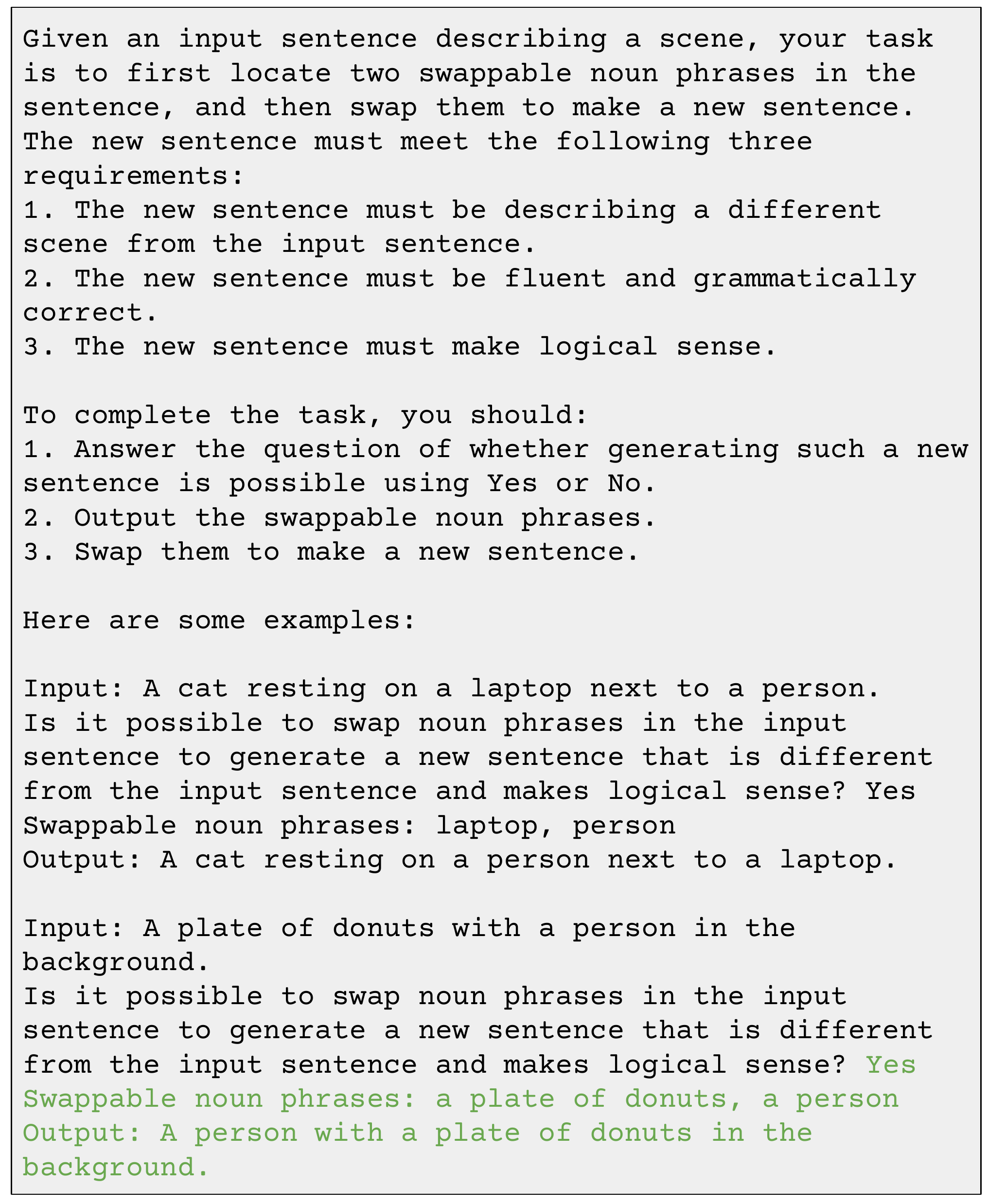}
  \caption{\swapobj.}
\end{subfigure}
\begin{subfigure}{.4\textwidth}
  \centering
  \includegraphics[width=\linewidth]{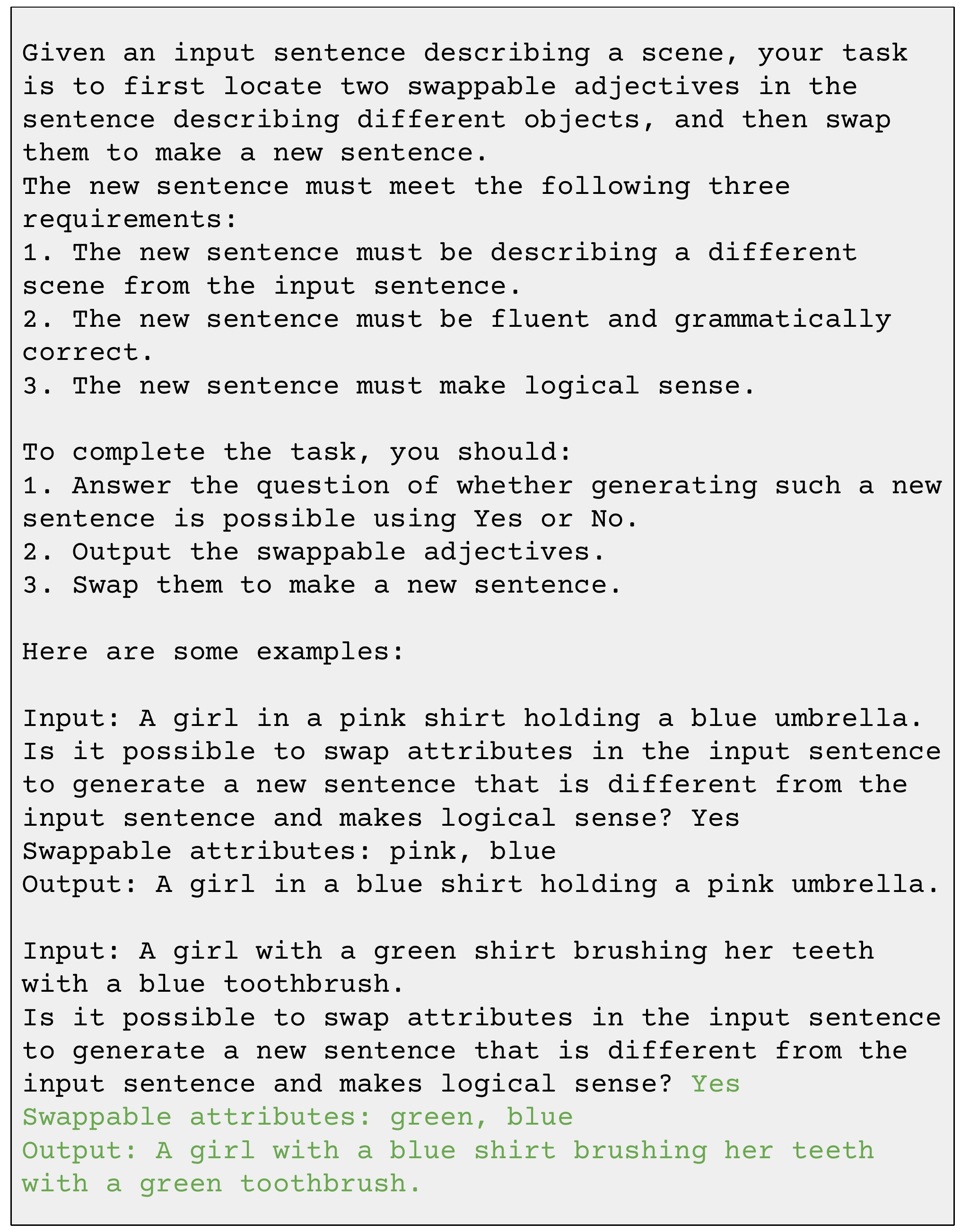}
  \caption{\swapatt.}
\end{subfigure}

\caption{Example prompt templates (black) and outputs (green) from ChatGPT for \swap\ hard negatives.}
\label{fig:prompt-swap}
\end{figure}

\begin{figure}[t]
\centering
\begin{subfigure}{.4\textwidth}
  \centering
  \includegraphics[width=\linewidth]{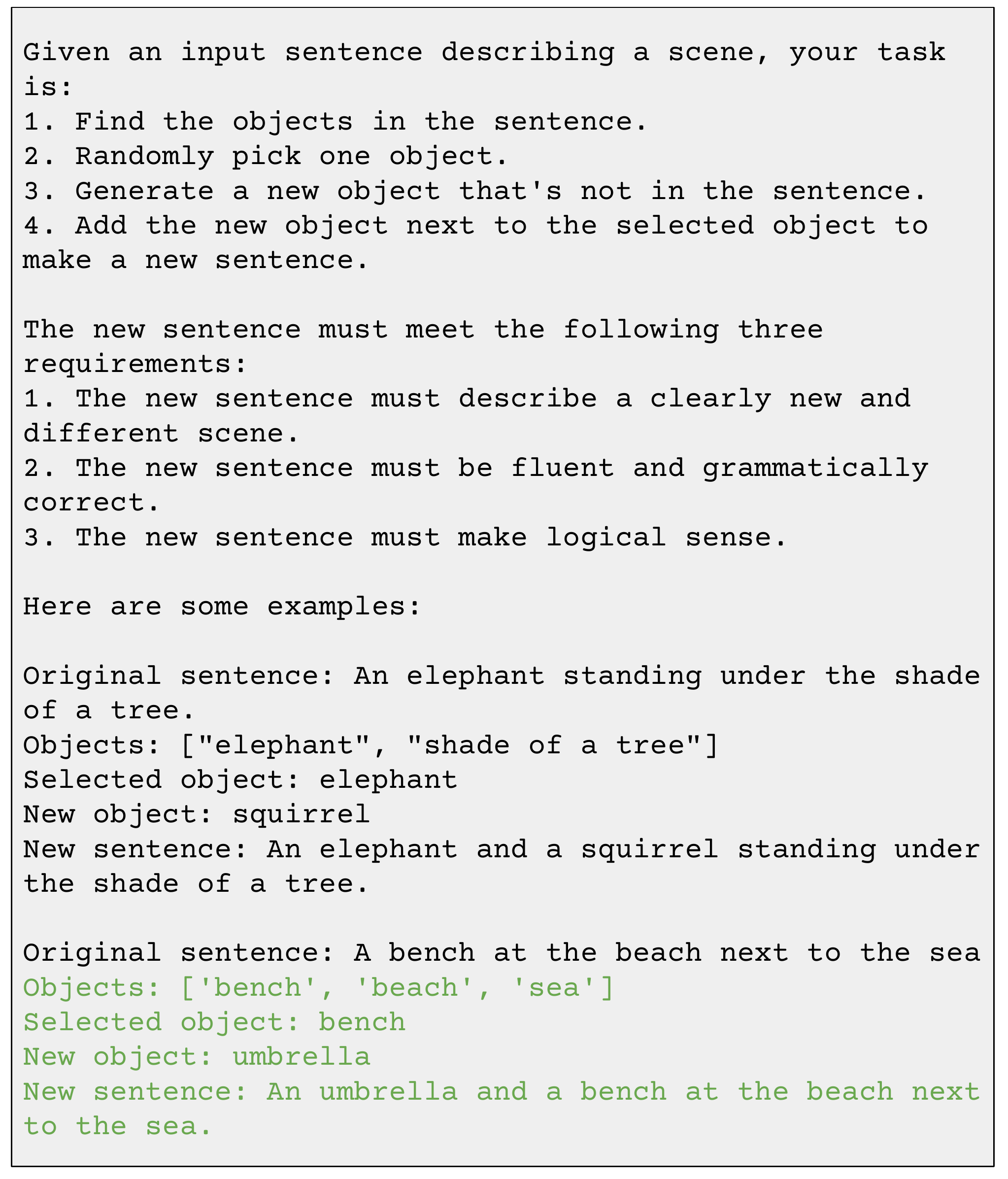}
  \caption{\addobj.}
\end{subfigure}
\begin{subfigure}{.4\textwidth}
  \centering
  \includegraphics[width=\linewidth]{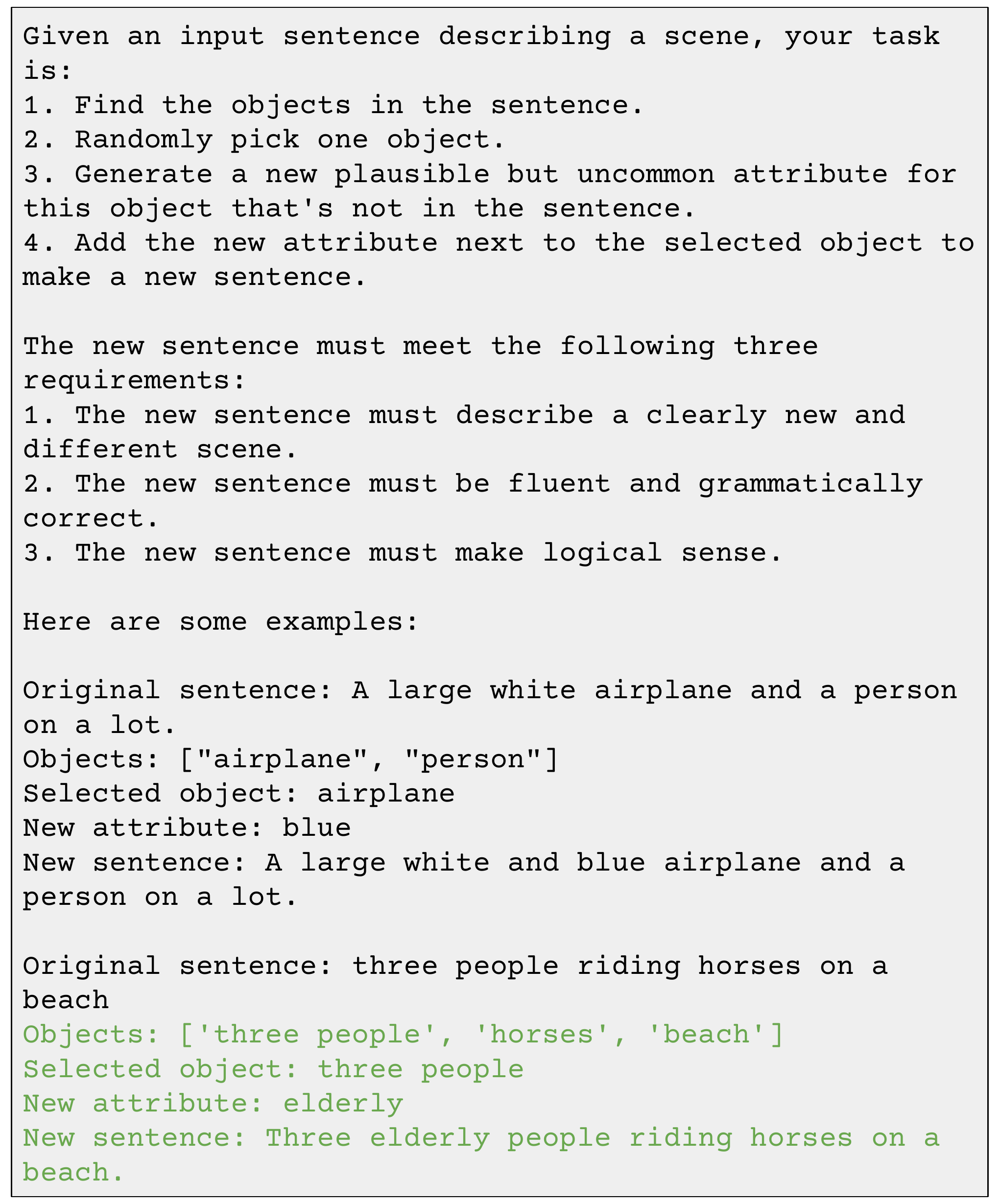}
  \caption{\addatt.}
\end{subfigure}
\caption{Example prompt templates (black) and outputs (green) from ChatGPT for \add\ hard negatives.}
\label{fig:prompt-add}
\end{figure}

\textbf{Generating \replace\ hard negatives.} 
To best leverage ChatGPT's capabilities, we devise a three-step workflow to generate \replace\ hard negatives: (1) We prompt ChatGPT in locating the desired atomic concepts (\eg, objects) in the sentence; (2) We prompt ChatGPT to generate a new concept to replace a randomly selected old concept; (3) We let ChatGPT compose a new sentence by replacing the old concept with the new one.
For steps (1) and (3), we prompt ChatGPT with a temperature of $0.0$ to get stable outputs. For step (2), however, we diversify the outputs by prompting ChatGPT with a higher temperature of $1.5$.
With this design, we are able to generate diverse \replace\ hard negatives.
Figure~\ref{fig:prompt-replace} shows the example templates and outputs for \replace\ hard negatives.

\textbf{Generating \swap\ hard negatives.} To generate \textsc{swap} hard negatives, which do not require any new concepts, we simply prompt ChatGPT once with a temperature of 0.0. Unlike \replace, \swap\ hard negatives are only possible when there are at least two atomic concepts of the same category, \ie, either object or attribute. Thus, our prompt first queries ChatGPT whether it is possible to swap two atomic concepts in the input sentence to generate a new description. Only if the answer is yes, will ChatGPT then proceed to identify two swappable concepts and compose the corresponding new sentence by swapping the two concepts.
Figure~\ref{fig:prompt-swap} shows the example templates and outputs for \swap\ hard negatives.

\textbf{Generating \add\ hard negatives.} Similar to the \replace, we also employ a three-step prompting procedure to generate \add\ hard negatives. The only difference in the procedure is that we prompt ChatGPT to add the generated new concept to the original caption, instead of using it to replace an old concept.
Figure~\ref{fig:prompt-add} shows the example templates and outputs for \add\ hard negatives.

\subsection{Adversarial refinement}
\label{app:refine}
We detail the adversarial refinement procedure below.
Given a text model $M$, we denote its output score for the positive and negative caption of $i$-th image as $M(p_{i})$ and $M(n_{i})$. If $M(p_{i})>M(n_{i})$, then the model could identify the correct caption for the $i$-th image without referring to it. For a test set to be unattackable given the text model $M$, the expectation of $M$'s identifying the correct caption should be as close to random guess as possible; in particular, we hope that $E_{i}[M(p_{i})>M(n_{i})]=0.5$. To achieve this for both the grammar model $M_1$ and plausibility model $M_2$, we first calculate the score difference $g^{(1)}_i=M_{1}(p_{i})-M_{1}(n_{i})$ and  $g^{(2)}_i=M_{2}(p_{i})-M_{2}(n_{i})$, where the range of both $g^{(1)}$ and $g^{(2)}$ is $[-1, 1]$. Then we split the 2D space of the joint range of  $g^{(1)}$ and $g^{(2)}$ into $100\times 100$ equal grids, and for each pair of symmetric grids, \eg, $\{(g^{(1)}, g^{(2)})|g^{(1)}\in(0.02, 0.04], g^{(2)}\in(-0.04, 0.06]\}$ and $\{(g^{(1)}, g^{(2)})|g^{(1)}\in(-0.02, -0.04], g^{(2)}\in(0.04, -0.06]\}$, we preserve the same number of data for both grids, therefore we ensure that for the resultant set, $E_{i}[M_{1}(p_{i})>M_{1}(n_{i})]=0.5$ and $E_{i}[M_{2}(p_{i})>M_{2}(n_{i})]=0.5$.

\subsection{Dataset information}
We host \method on Github~\footnote{\url{https://github.com/RAIVNLab/sugar-crepe}}.
The data card~\cite{pushkarna2022data} for \method, containing detailed dataset documentation, is available at the dataset repository~\footnote{\url{https://github.com/RAIVNLab/sugar-crepe/blob/main/data_card.pdf}}.
We provide a summary below.

\textbf{Dataset documentation.}
\method is a benchmark for faithful vision-language compositionality evaluation. Given an image, a model is required to select the positive text that correctly describes the image, against another hard negative text distractor that differs from the positive text only by small compositional changes. Each example consists of three fields:
\begin{itemize}
    \item \texttt{filename}: The id to an image
    \item \texttt{caption}: Positive text correctly describing the image
    \item \texttt{negative\_caption}: Hard negative text incorrectly describing the image
\end{itemize}

\textbf{Maintenance plan.}
We are committed to maintain the dataset to address any technical issues. We actively monitor issues in the repository.

\textbf{Licensing.}
We license our work using MIT License~\footnote{\url{https://github.com/RAIVNLab/sugar-crepe/blob/main/LICENSE}}.
All the source data we use is publicly released by prior work~\cite{lin2014microsoft}.

\textbf{Author statement.}
We the authors will bear all responsibility in case of violation of rights.

\section{Detailed evaluation results}

\subsection{Full evaluation results on existing benchmarks}
\label{app:eval}

We provide the full evaluation results over $17$ pretrained CLIP models as well as $2$ text-only models, Vera~\cite{liu2023vera} and the Grammar model~\cite{morris2020textattack}, on existing compositionality benchmarks in Table~\ref{tab:eval_existing}. We see that the text-only models, arguably without any vision-language compositionality, outperform most of the pretrained CLIP models, achieving state-of-the-art performances on many benchmark tasks. This implies that current benchmarks fail to faithfully reflect a model's vision-language compositionality.

\begin{table*}[h]
  \centering
  \small
  \caption{Blind models (\ie, Vera and Grammar model) outperform all $17$ existing pretrained CLIP models on nearly all existing benchmark tasks. This implies that current benchmarks fail to faithfully measure a model's vision-language compositionality.}
  \scalebox{0.55}{
    \begin{tabular}{ll cccccccccc} 
    \toprule
     & & \multicolumn{3}{c}{\bf CREPE} &\multicolumn{4}{c}{\bf ARO} &\multicolumn{3}{c}{\bf VL-Checklist}\\ 
     \cmidrule(lr){3-5}\cmidrule(lr){6-9}\cmidrule(lr){10-12}
     
    \textbf{Source} & \textbf{Model} &   \textbf{Atomic} & \textbf{Swap} & \textbf{Negate} & \textbf{VG-Relation} & \textbf{VG-Attribution} & \textbf{COCO-Order} & \textbf{Flickr30K-Order} & \textbf{Object} & \textbf{Attribute} & \textbf{Relation}  \\ 
    \midrule

\multirow{2}{*}{Text-only model} &  Vera~\cite{liu2023vera}   & 43.70 & 70.80 & 66.15 & 61.71 & 82.59 & 59.81 & 63.52 & 82.48 & 73.99 & 85.72\\
& Grammar~\cite{morris2020textattack} & 18.15 & 50.88 & 9.77 & 59.55 & 58.38 & 74.33 & 76.26 & 57.95 & 52.35 & 68.50\\

\midrule\midrule
   
\multirow{7}{*}{OpenAI~\cite{RadfordKHRGASAM21}} 
& RN50 & 26.47  & 28.32 & 31.25& 53.87 & 63.37 & 44.89 & 52.46 & 86.85 & 68.30 & 75.95\\
& RN101 & 27.63  & 32.74 & 12.50 & 52.43 & 62.93 & 29.86 & 39.34 & 86.44 & 67.93 & 71.75\\
& RN50x4 & 26.24 & 28.32 & 9.51 & 51.59 & 62.27 & 29.39 & 34.56 & 87.23 & 68.74 & 73.81\\
& ViT-B-32 & 22.31 & 26.55 & 28.78 & 51.12 & 61.33 & 37.14 & 47.18 & 87.00 & 68.80 & 77.04\\
& RN50x16 & 26.36 & 29.65 & 9.38 & 52.13 & 62.71 & 29.95 & 34.26 & 86.95 & 69.34 & 76.83\\
& RN50x64 & 26.82 & 30.09 & 23.57 & 51.00 & 62.56 & 40.54 & 46.74 & 87.71 & 68.61 & 74.97\\
& ViT-L-14 & 26.36 & 25.66 & 24.74 & 53.34 & 61.50 & 36.11 & 45.08 & 87.86 & 68.27 & 75.89\\

\midrule

\multirow{6}{*}{LAION~\cite{laion5b}}  & ViT-H-14 & 23.70 & 25.22 & 16.54 & 50.33 & 62.93 & 25.79 & 30.96 & 85.39 & 68.46 & 71.13\\
& ViT-g-14 & 23.70 & 24.78 & 20.70 & 51.60 & 61.20 & 25.59 & 30.10 & 86.07 & 69.43 & 71.03\\
& ViT-bigG-14 & 23.58 & 24.78 & 17.97 & 51.61 & 61.89 & 25.24 & 30.22 & 84.66 & 67.80 & 66.48\\
& roberta-ViT-B-32 & 22.66 & 21.24 & 20.31 & 47.46 & 62.00 & 24.77 & 30.76 & 85.71 & 68.82 & 65.90\\
& xlm-roberta-base-ViT-B-32 & 21.16 & 20.80 & 12.76 & 47.93 & 59.73 & 23.85 & 30.32 & 86.06 & 70.41 & 63.01\\
& xlm-roberta-large-ViT-H-14 & 24.16 & 23.89 & 20.05 & 46.14 & 57.84 & 26.05 & 31.00 & 87.89 & 70.25 & 63.89\\

\midrule

\multirow{4}{*}{DataComp~\cite{gadre2023datacomp}}  
& \texttt{small:}ViT-B-32 & 13.64 & 27.88 & 14.84 & 50.83 & 50.17 & 13.35 & 14.02 & 68.72 & 58.80 & 57.00\\
& \texttt{medium:}ViT-B-32 & 16.42 & 20.35 & 11.33 & 50.45 & 54.04 & 16.44 & 16.26 & 78.43 & 63.53 & 62.94\\
& \texttt{large:}ViT-B-16 & 18.15 & 17.26 & 17.06 & 48.82 & 53.21 & 21.49 & 26.44 & 84.73 & 65.72 & 64.81\\
& \texttt{x-large:}ViT-L-14 & 21.62 & 22.57 & 16.28 & 48.54 & 60.03 & 23.19 & 29.52 & 86.66 & 67.01 & 67.93\\

\bottomrule 
\end{tabular}
}
  
  \label{tab:eval_existing}

\end{table*}

\subsection{\method human evaluation}
\label{app:sugar}
To compare the quality of the hard negatives generated in \method to those in current benchmarks (\ie, \hncoco), we randomly sample $100$ examples for each of the hard negative types: \replace, \swap, and \negate\ / \add.
Each example is organized to consist of (1) the original positive text, (2) its hard negative in \hncoco, and (3) its hard negative in \method. For each example, a human user rates whether the hard negative in \hncoco\ or that in \method is better (or tie) in terms of commonsense and grammatical correctness, respectively. Note that we compare \negate\ in \hncoco\ to \add\ in \method, as both hard negatives are intended to probe a model's understanding of the \textit{existence or not} of an atomic concept. Table~\ref{tab:sugar_crepe_eval_quant_human} shows that hard negatives in \method are much more sensical and fluent than that in \hncoco\ across all three different types. For instance, \method has $68\%$ more sensical and $46\%$ more fluent hard negatives than \hncoco\ on \swap.

\begin{table*}[h]
  \centering
  \small
  \caption{Human evaluation results on the comparisons between hard negatives in \hncoco\ and \method. We report the counts (out of $100$ sampled examples) that the human user considers better or tie, w.r.t. both commonsense and grammatical correctness.}
  \scalebox{1.0}{
    \begin{tabular}{ll ccc} 
    \toprule

    & & \multicolumn{3}{c}{Human counts of better examples} \\
    \cmidrule{3-5} 
    Hard-negative Type & Evaluation & \hncoco & \method\ & Tie \\
    \midrule
    
    \multirow{2}{*}{\replace} & Commonsense & 11 & 29 & 60 \\
    & Grammar  & 4 & 33 & 63 \\
    
    \midrule

    \multirow{2}{*}{\swap}  & Commonsense  & 4 & 68 & 28 \\
    & Grammar  & 4 & 46 & 50 \\
    
    \midrule

    \multirow{2}{*}{\negate\ / \add}  & Commonsense  & 1 & 26 & 73 \\
    & Grammar  & 1 & 35 & 64 \\
    
    \bottomrule 

\end{tabular}
}
  
  \label{tab:sugar_crepe_eval_quant_human}

\end{table*}

\end{document}